\documentclass{article}

\usepackage[preprint]{neurips_2025}

\usepackage[utf8]{inputenc} 
\usepackage[T1]{fontenc}    
\usepackage[colorlinks=true,
            linkcolor=linkcolor,
            citecolor=citecolor,
            urlcolor=cyan,
            breaklinks=true,
            pdfstartview=FitH]{hyperref}
\usepackage{url}            
\usepackage{booktabs}       
\usepackage{amsfonts}       
\usepackage{nicefrac}       
\usepackage{microtype}      
\usepackage[table]{xcolor}         
\definecolor{citecolor}{HTML}{0071BC}
\definecolor{linkcolor}{HTML}{0071BC}

\usepackage{multirow}
\usepackage{amsmath}
\usepackage{arydshln}
\usepackage{graphicx}
\usepackage[most]{tcolorbox}
\usepackage{amssymb}
\usepackage{enumitem}
\usepackage{listings}
\usepackage{cleveref}
\usepackage{arydshln}

\usepackage{minitoc}

\noptcrule

\usepackage{array}
\newcolumntype{P}[1]{>{\raggedright\arraybackslash}p{#1}} 
\newcolumntype{C}[1]{>{\centering\arraybackslash}p{#1}}   
\newcolumntype{R}[1]{>{\raggedleft\arraybackslash}p{#1}}  

\newcommand*\samethanks[1][\value{footnote}]{\footnotemark[#1]}

\makeatletter
\AtBeginDocument{%
  \renewcommand{\sectionautorefname}{\S\@gobble}

  \renewcommand{\subsectionautorefname}{\S\@gobble}  
  \renewcommand{\subsubsectionautorefname}{\S\@gobble} 
}
\makeatother

\definecolor{lightgray}{gray}{0.95}
\lstdefinestyle{promptstyle}{
  backgroundcolor=\color{lightgray},
  basicstyle=\ttfamily\small,
  breaklines=true,
  frame=single,
  rulecolor=\color{gray},
  showstringspaces=false,
  columns=fullflexible
}

\title{Think or Not? Exploring Thinking Efficiency in Large Reasoning Models via an Information-Theoretic Lens}

\author{Xixian Yong$^1$ \quad 
Xiao Zhou$^{123}$\thanks{Corresponding authors.} \quad 
Yingying Zhang$^4$ \quad 
Jinlin Li$^1$ \\
\textbf{Yefeng Zheng}$^5$ \quad 
\textbf{Xian Wu}$^{4}$\samethanks \quad 
\\
\textsuperscript{1}Gaoling School of Artificial Intelligence, Renmin University of China \\
\textsuperscript{2}Beijing Key Laboratory of Research on Large Models and Intelligent Governance \\
\textsuperscript{3}Engineering Research Center of Next-Generation Intelligent Search and Recommendation, MOE \\
\textsuperscript{4}Tencent Jarvis Lab \quad 
\textsuperscript{5}Medical Artificial Intelligence Lab, Westlake University \\
\texttt{\{xixianyong, xiaozhou\}@ruc.edu.cn} \quad 
\texttt{kevinxwu@tencent.com}
}

\begin{document}
\doparttoc
\faketableofcontents
\maketitle

\begin{abstract}
The recent rise of Large Reasoning Models (LRMs) has significantly improved multi-step reasoning performance, but often at the cost of generating excessively long reasoning chains. This paper revisits the efficiency of such reasoning processes through an information-theoretic lens, revealing a fundamental trade-off between reasoning length and semantic efficiency. We propose two metrics—\textit{InfoBias} and \textit{InfoGain}—to quantify divergence from ideal reasoning paths and stepwise information contribution, respectively. Empirical analyses show that longer reasoning chains tend to exhibit higher information bias and diminishing information gain, especially for incorrect answers. Motivated by these findings, we introduce an \textbf{entropy-based Adaptive Think} strategy that dynamically halts reasoning once confidence is sufficiently high, improving efficiency while maintaining competitive accuracy. Compared to the Vanilla Think approach (default mode), our strategy yields a 1.10\% improvement in average accuracy and a 50.80\% reduction in token usage on QwQ-32B across six benchmark tasks spanning diverse reasoning types and difficulty levels, demonstrating superior efficiency and reasoning performance. These results underscore the promise of entropy-based methods for enhancing both accuracy and cost-effiiciency in large language model deployment. Code and data are available at \url{https://github.com/chicosirius/think-or-not}.
\end{abstract}

\section{Introduction}
With the paradigm of Large Language Models (LLMs)~\citep{brown2020language} extending from training-time scaling~\citep{kaplan2020scaling} to test-time scaling~\citep{muennighoff2025s1}, the emergence of Large Reasoning Models (LRMs)~\citep{li2025system}—such as OpenAI’s o1~\citep{o1}, Deepseek’s R1~\citep{guo2025deepseek}, and QwQ-32B~\citep{qwq32b}—has significantly advanced the frontier of model reasoning capabilities. However, we observe a noteworthy trend: in pursuit of better performance, these models increasingly rely on lengthy Chain-of-Thought (CoT)~\citep{wei2022chain} reasoning, leading to quadratic growth in computational complexity. This prolonged internal or external “deep thinking” process contradicts the principle of cognitive economy observed in human reasoning, thereby undermining the efficiency of LRMs in practical applications~\citep{su2025between}.

\begin{figure}[t]
\centering
\includegraphics[width=\textwidth]{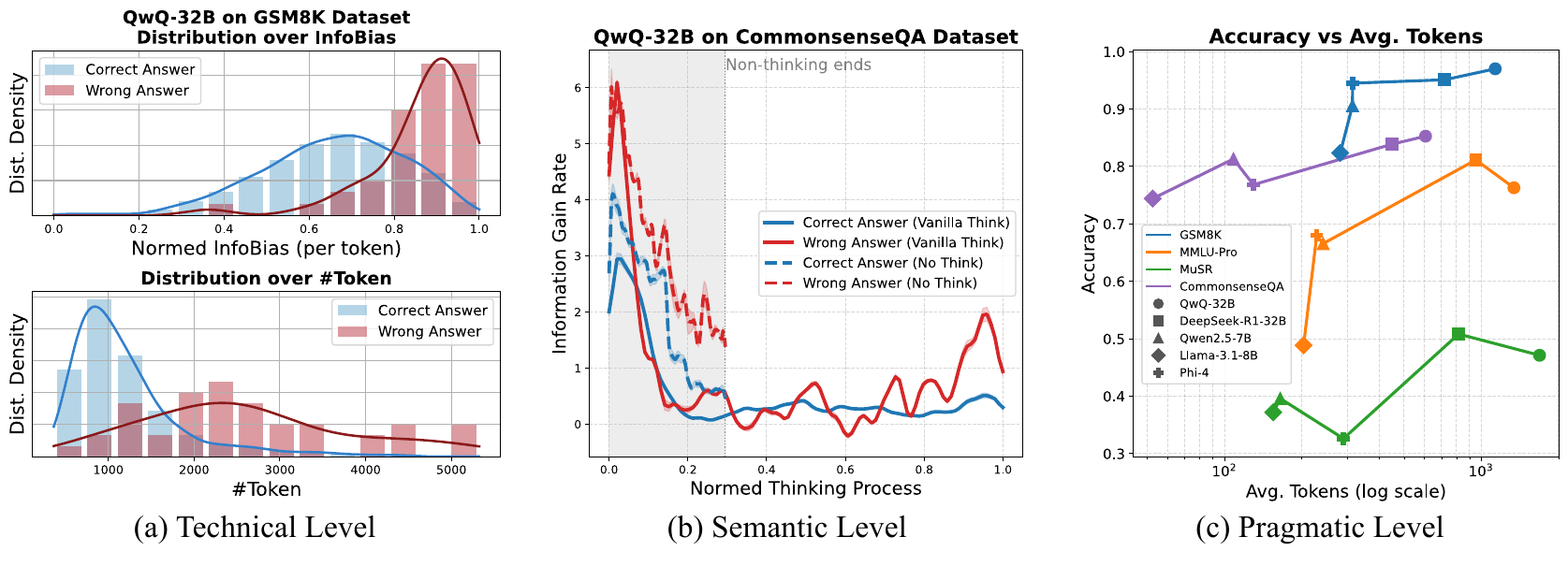}
\caption{
\textbf{Understanding thinking inefficiency via Shannon \& Weaver’s Communication Model.} (a) Technical Level: On the GSM8K dataset, incorrect answers exhibit higher InfoBias and longer token lengths, suggesting that longer reasoning does not necessarily lead to better outcomes. (b) Semantic Level: The InfoGain rate shows a nonlinear decline as the thinking progresses, indicating diminishing contribution to entropy reduction over the final answer space. (c) Pragmatic Level: Results across various models and benchmarks show longer reasoning yields diminishing returns and may even reduce final accuracy. Detailed calculation methods and analysis are provided in \autoref{sec.quantifying}.
}
\label{fig:info-level}
\end{figure}

Inspired by Shannon’s three-level model of communication~\citep{shannon1948mathematical}, we revisit the phenomenon of excessively long reasoning chains in contemporary LRMs. \textbf{At the technical level}, extending the reasoning chain can be interpreted as injecting redundant bits into a noisy channel to enhance robustness against perturbations~\citep{min2021noisy}. However, once the reasoning length exceeds the model’s reasoning capacity—an analogue to channel capacity—additional redundancy ceases to improve accuracy and instead induces error accumulation and semantic drift (\autoref{fig:info-level}(a)). \textbf{At the semantic level}, as the number of reasoning steps increases, the information gain per step rapidly diminishes; excessive reasoning contributes little to uncertainty reduction and may even introduce semantic noise, revealing inefficiencies in the mapping between symbols and meanings (\autoref{fig:info-level}(b)). \textbf{At the pragmatic level}, while longer reasoning chains may improve interpretability, they impose higher computational and cognitive costs, often yielding diminishing returns~\citep{sprague2024cot}—or even performance degradation—on various tasks (\autoref{fig:info-level}(c)).

This multi-level inefficiency highlights a central contradiction in the current LRM reasoning paradigm: substantial compute investments do not consistently translate into semantic efficiency or downstream performance gains. Motivated by this insight, we pose a core question: \textbf{Can we optimize the reasoning patterns of LRMs to substantially shorten reasoning chains while maintaining performance across diverse reasoning tasks?}

To quantitatively assess the efficiency of a model’s reasoning process, we adopt an information-theoretic perspective and conduct in-depth analysis at two levels: (i) the response-level information bias, where we compute the mutual information between the model’s generated response and the ground-truth reasoning trajectory to estimate InfoBias, capturing the overall semantic alignment across the full reasoning output (\autoref{sec.infobias}); (ii) the step-level information gain, where we quantify InfoGain as the entropy reduction over the answer distribution induced by each reasoning step, reflecting how much new information is introduced at each stage of the reasoning process (\autoref{sec.infogain}). Our empirical experiments (\autoref{sec.3.4}) reveal a significant, nonlinear positive correlation between reasoning length and InfoBias. Notably, incorrect answers consistently exhibit higher InfoBias, and the lengths of their generated responses are often biased toward longer reasoning chains. Furthermore, step-wise analysis indicates that models often possess a degree of intuitive confidence about the correct answer even before any explicit reasoning occurs. As reasoning unfolds, the InfoGain over the answer space and the model’s confidence in the correct answer evolve differently across various types of reasoning tasks. While non-reasoning modes yield higher InfoGain per step, they typically result in lower overall confidence in the final answer compared to their reasoning-enabled counterparts.


Based on these analyses, we propose an \textbf{entropy-based Adaptive Think} strategy that dynamically halts reasoning once the model’s confidence—quantified via entropy over the answer space—exceeds a tunable threshold (\autoref{sec.method}). We compare this approach against three alternative strategies: \textbf{Vanilla Think}, \textbf{No-Think}, and \textbf{Gated Think}. Extensive experiments (\autoref{sec.results}) are conducted across five language models and six benchmarks covering diverse types of reasoning tasks. Experimental results demonstrate that our Adaptive Think improves both accuracy and reasoning efficiency across mathematical, factual, logical, and commonsense reasoning tasks. On two math benchmarks of varying difficulty, our method reduces token usage by 58.78\% while preserving— and slightly improving— accuracy (average +0.95\%). Beyond math, it boosts model accuracy by an average of 0.38\% and reduces token usage by 42.39\% across non-mathematical reasoning tasks. Finally, we conduct an in-depth analysis (\autoref{sec.analysis}) of when and how much reasoning a model should perform.

\section{Related Work}
\paragraph{Information-Theoretic Perspectives}
Information theory has long served as a lens for analyzing machine learning systems, particularly in studying generalization bounds~\citep{russo2016controlling,xu2017information} and understanding learning objectives~\citep{slonim2002unsupervised}. Recent work extends these ideas to LLMs, using entropy-based measures to evaluate reasoning reliability~\citep{ton2024understanding,gan2025rethinking}. Semantic entropy, in particular, has been proposed as a tool for detecting hallucinations by measuring variability in meaning across generations~\citep{farquhar2024detecting}, and can be efficiently estimated using hidden states alone~\citep{kossen2024semantic}. Other approaches use entropy to identify reasoning failures in multi-step generation without requiring supervision~\citep{ali2025entropy}.

\paragraph{Adaptive and Efficient Reasoning}
Efficiency in LLMs is an active area of research, with methods that adapt step counts based on task difficulty, confidence, or resource constraints~\citep{han2024token,pan2024dynathink,shen2025dast}. Early exit mechanisms and dynamic token allocation~\citep{yang2025dynamic,qu2025optimizing} aim to reduce unnecessary computation, while approaches such as elastic CoT and multi-scale reasoning seek to better align model capacity with problem complexity~\citep{ma2025cot,kirchner2024prover}. Studies have shown that longer CoT do not always improve performance~\citep{wu2025more,yang2025towards}, and in some cases can lead to overthinking, particularly in high-capacity models~\citep{chen2024not}. This has led to interest in minimal or even implicit reasoning strategies~\citep{ma2025reasoning,sui2025stop}, emphasizing the need for more nuanced reasoning strategies and adaptive control over reasoning depth.

\section{Quantifying Thinking Efficiency}
\label{sec.quantifying}
This section introduces a formal framework to measure reasoning efficiency by segmenting the thought process, analyzing divergence from ideal reasoning paths, and computing stepwise information gains.

\subsection{Semantic Segmentation of Thinking Processes}
Human reasoning typically unfolds in discrete, sequential steps. The means–ends analysis framework~\citep{simon1971human} views problem solving as a series of goal-subgoal transitions, each representing a cognitive operation. Similarly, ACT-R~\citep{anderson1997act, whitehill2013understanding} models reasoning as rule-based production sequences, while dual-process theory~\citep{kahneman2011thinking, evans2003two} characterizes “System 2” reasoning as deliberate and decomposable. Collectively, these theories motivate modeling reasoning as a structured sequence of semantically meaningful steps.

Accordingly, we segment a model’s output reasoning path $S$ into discrete semantic units $S = \{s_1, s_2, \cdots, s_n\}$, where each $s_i$ represents a minimal step that contributes semantically to the overall process. For example, "solving $2x + 5 = 15$" triggers steps as $s_1$: subtract $5$ from both sides → $s_2$: divide both sides by $2$ → $s_3$: solve for $x$. These segments serve as the atomic elements for downstream information-theoretic analysis. The segmentation can be performed based on syntactic cues (e.g., clause or sentence boundaries), manual annotation, or automated approaches such as LLM-assisted chunking. By operating at this granularity, we enable a finer analysis of how incremental reasoning steps influence uncertainty and information flow throughout the trajectory.

\subsection{Response-Level: Measuring Information Bias in Entire Trajectories}
\label{sec.infobias}
While $S$ captures the model’s observable reasoning path, we posit the existence of a latent, ideal trajectory $T=\{t_1, t_2, \cdots, t_m\}$ representing the correct reasoning steps for a given question $Q$. This ideal trajectory may correspond to a human-annotated, cognitively plausible reasoning path, or reflect implicit reasoning steps within the model itself~\citep{gan2025rethinking}, which may differ from its explicit outputs. To measure how closely the model’s reasoning aligns with this ground truth, we introduce information bias, a metric based on mutual information:
\begin{equation}
    {\rm InfoBias}(S,T) = -I(s_{1:n}, t_{1:m}) = H(s_{1:n}, t_{1:m}) - H(s_{1:n}) - H(t_{1:m}),
\end{equation}
where $I$ denotes mutual information and $H$ represents entropy. This discrepancy can be estimated via sampling, under the assumption that the generated reasoning trajectories $s$ and $t$ are two conditionally independent stochastic processes, and their joint distribution can be approximated through $N$ samples. Applying the KL-based estimation of mutual information~\citep{paninski2003estimation}, we derive the following upper bound on the information bias:
\begin{equation}
    |\hat{I}_N(S,T) - I(S,T)| \le \sqrt{\frac{2 \log (2/\delta)}{N}} + \mathcal{O}\left (\frac{1}{N}\right ),
\end{equation}
where $\delta$ denotes the confidence level. This bound guarantees that the empirical estimate $\hat{I}_N(S,T)$ converges to the true mutual information $I(S,T)$ as $N$ increases, establishing $\rm InfoBias$ as a statistically consistent metric. Crucially, this enables reliable estimation of the alignment between observable and latent reasoning trajectories using a finite number of sampled inference steps.

\subsection{Step-Level: Measuring Information Gain at Each Step}
\label{sec.infogain}
Beyond the trajectory as a whole, at the semantic level, we aim to quantify how each individual reasoning step contributes to answer inference. Efficient reasoning should progressively reduce uncertainty over the answer space~\citep{sui2025stop}. Given a set of candidate answers $A = \{a_1, a_2, \cdots, a_l\}$, we can compute the conditional entropy at step $i$:
\begin{equation}
    H_i = - \sum_{k=1}^l P(a_k|Q; s_{1:i}) \log P(a_k|Q; s_{1:i}),
\end{equation}
where $P(a_k|Q; s_{1:i})$ is estimated from the model's output probabilities. Specifically, we concatenate the given question $Q$, the model’s intermediate reasoning steps $s_{1:i}$, and the final answer prompt to form the input sequence (See {\color{citecolor}Appendix C.3} for details). The model’s predicted probability of the next token is then used as the basis for evaluation. The information gain at step $i$ is:
\begin{equation}
    \Delta I_i = H_{i-1} - H_i,
\end{equation}
which quantifies how much uncertainty is reduced by incorporating step $s_i$. This reflects the extent to which each reasoning step clarifies the answer distribution. We further define a \textbf{targeted information gain} with respect to the correct answer $c \in A$:
\begin{equation}
    \Delta I_i^c = - \log P(c|Q; s_{1:i}) - (- \log P(c|Q; s_{1:i-1})) = \log \frac{P(c|Q; s_{1:i-1})}{P(c|Q; s_{1:i})},
\end{equation}
capturing how each step influences the model’s confidence in the correct option. Together, $\Delta I_i$ and $\Delta I_i^c$ reveal fine-grained reasoning efficiency, highlighting impactful steps toward the correct answer.

\subsection{Empirical Evaluation and Insights}
\label{sec.3.4}
We empirically validate the methods proposed in \autoref{sec.infobias} and \autoref{sec.infogain}, which respectively target the response-level relationship between reasoning length and InfoBias, and the step-level impact of individual reasoning steps on InfoGain. These analyses aim to assess the effectiveness of the information-theoretic metrics in capturing the dynamics and quality of reasoning exhibited by LLMs.

\subsubsection{InfoBias and the Risks of Overgeneration}

\begin{figure}[ht]
\centering
\includegraphics[width=\textwidth]{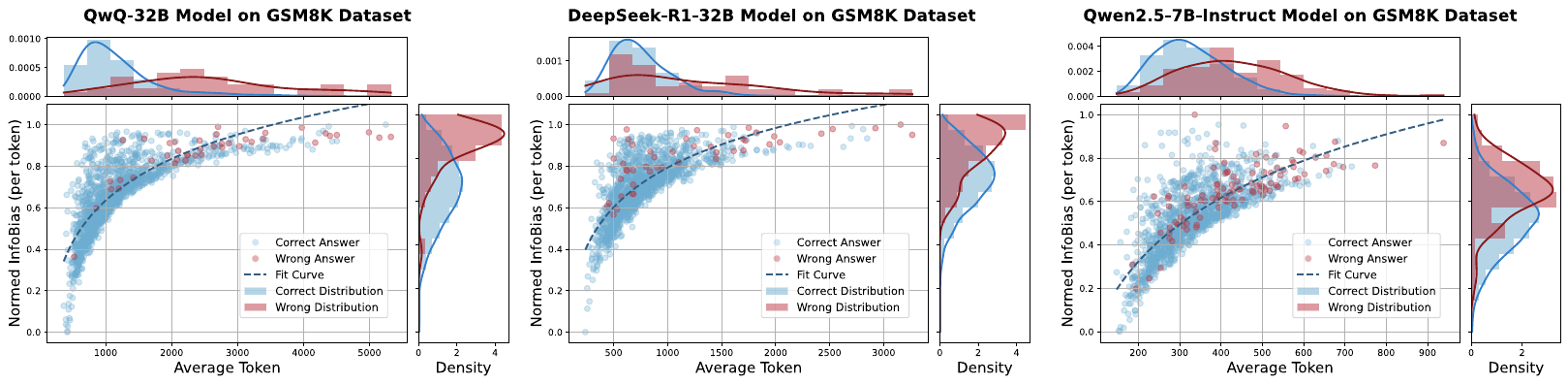}
\caption{
\textbf{Normalized InfoBias per token as a function of average reasoning length for different models on the GSM8K dataset.} Blue and red points represent instances with correct and incorrect answers, respectively, with density estimates of tokens and InfoBias shown on the top and right.
}
\label{fig:infobias}
\end{figure}
To examine the relationship between response length and semantic deviation, we compute InfoBias over samples drawn from both model generations and reformulated ground-truth rationales (see {\color{citecolor}Appendix C.2} for details). Results on GSM8K (\autoref{fig:infobias}) reveal two key observations:

\paragraph{Findings 1: Cumulative InfoBias with Increased Reasoning Length.}
We observe a consistent monotonic trend: longer reasoning chains tend to accumulate deviation from the correct reasoning path, suggesting that additional tokens often introduce noise rather than refinement. This pattern holds for both reasoning and non-reasoning models (see {\color{citecolor}Appendix D.1} for more results). There is no sign of InfoBias saturation or decline—even strong models exhibit rising bias, implying that simply generating more tokens does not guarantee improved alignment or correctness.

\textbf{Findings 2: Incorrect answers exhibit higher InfoBias and more variable response length.}
A pronounced separation is observed between correct and incorrect samples: incorrect answers show higher InfoBias and slightly longer reasoning chains, indicating that extended reasoning amplifies rather than corrects misalignment. Moreover, the length distribution of incorrect answers is broader, indicating greater variability and instability in how models diverge from the correct reasoning path.

\subsubsection{InfoGain and Step-Level Reasoning Quality}
\label{sec.infogain-cal}
\begin{figure}[ht]
\centering
\includegraphics[width=\textwidth]{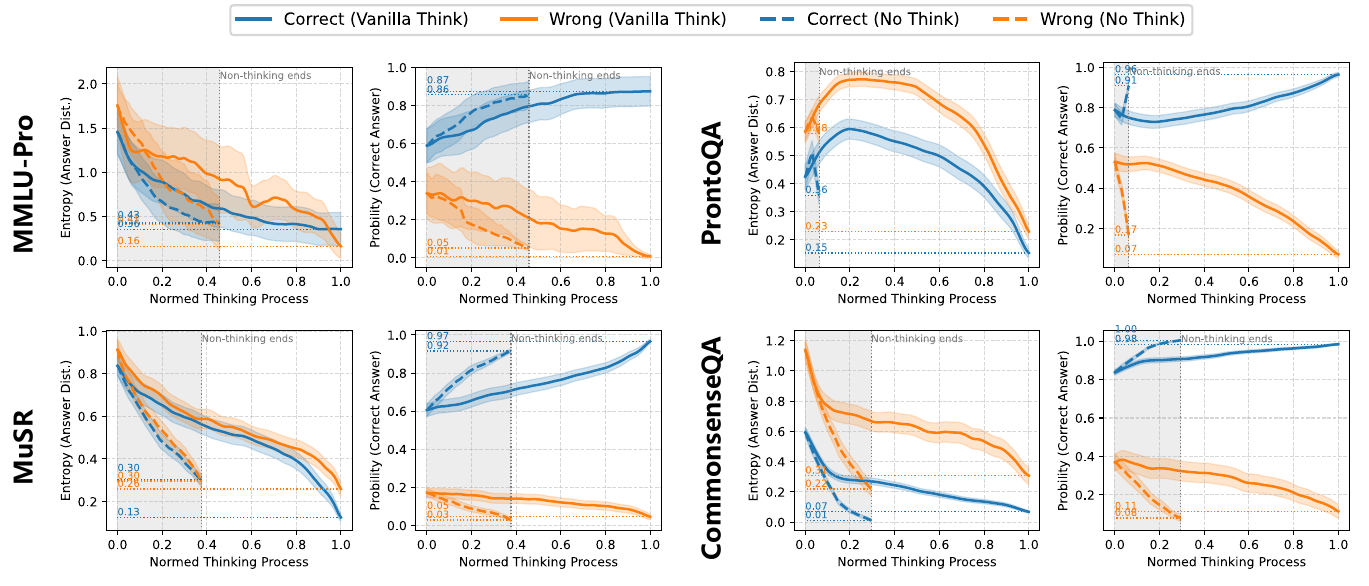}
\caption{
\textbf{Uncertainty dynamics across different reasoning benchmarks for QwQ-32B.} Each set includes two subplots: (1) entropy of the answer distribution vs. normalized reasoning steps, and (2) model-predicted probability of the correct answer over the same steps. Blue/orange lines denote correct/incorrect predictions; solid/dashed lines correspond to Vanilla Think and No-Think. Shaded areas mark the average token proportion used in No-Think mode. Step-wise analysis shows that models often exhibit early intuitive confidence in correct answers, even before reasoning starts. As reasoning unfolds, uncertainty decreases and confidence grows in task-specific ways.
}
\label{fig:infogain}
\end{figure}

We next turn to the dynamics of reasoning steps. By segmenting rationales into paragraph-level units and measuring per-step InfoGain, we analyze how entropy and confidence evolve during inference across multiple benchmarks (\autoref{fig:infogain}). Based on further analysis, we draw the following findings.


\textbf{Findings 3: Reasoning Steps Consistently Reduce Uncertainty.}
We observe that reasoning traces leading to correct answers consistently exhibit a reduction in entropy over the answer space and a corresponding increase in confidence for the correct choice. This supports the notion that effective reasoning incrementally filters uncertainty and sharpens prediction. Moreover, while No-Think mode achieves higher information efficiency per step—rapidly lowering entropy—it typically converges to lower final confidence, limiting its reliability. By comparison,Vanilla Think involves longer and less efficient reasoning chains in terms of information gain per step, but ultimately yields more confident and accurate predictions, underscoring a trade-off between efficiency and robustness in reasoning.

\textbf{Findings 4: Reasoning Models Exhibit Initial Intuition.}
Even before reasoning begins (step = 0), samples that eventually lead to correct answers already show lower entropy and higher confidence. This indicates that the models possess an initial bias or “intuitive prior” toward the correct answer even before engaging in multi-step reasoning. This effect is especially pronounced in knowledge-intensive tasks like MMLU-Pro and CommonsenseQA, suggesting that LRMs often start with strong inductive biases toward the correct choice, possibly due to extensive prior exposure during training.

\textbf{Findings 5: Task-Specific Reasoning Dynamics.}
In CommonsenseQA, entropy drops rapidly at the early stages, suggesting that commonsense questions can often be resolved with minimal reasoning. Notably, No-Think mode yields higher final confidence than Vanilla Think, implying that the latter's intermediate reasoning steps may be redundant or inefficient. Meanwhile, MMLU-Pro and MuSR show smooth and monotonic entropy separation between correct and incorrect samples, reflecting tasks where gradual semantic integration is beneficial. In contrast, ProntoQA exhibits a non-monotonic pattern—entropy first rises, then falls—which may result from its binary format: early steps broaden the hypothesis space and reduce overconfidence before eventual convergence. Overall, these dynamics reflect how the task’s type influence the utility of the reasoning process.

These findings highlight the potential of entropy-based signals as proxies for monitoring and controlling reasoning in LRMs. The steady accumulation of InfoBias with longer reasoning suggests that unregulated generation often leads to semantic drift, while InfoGain trends reveal diminishing returns from extended reasoning. Early confidence signals also suggest that further reasoning is often unnecessary. These insights motivate our approach: adaptively modulating reasoning depth based on entropy, allowing models to think when needed and stop when additional steps offer little value.

\section{Entropy-Based Adaptive Thinking}
\label{sec.method}
\begin{figure}[ht]
\centering
\includegraphics[width=\textwidth]{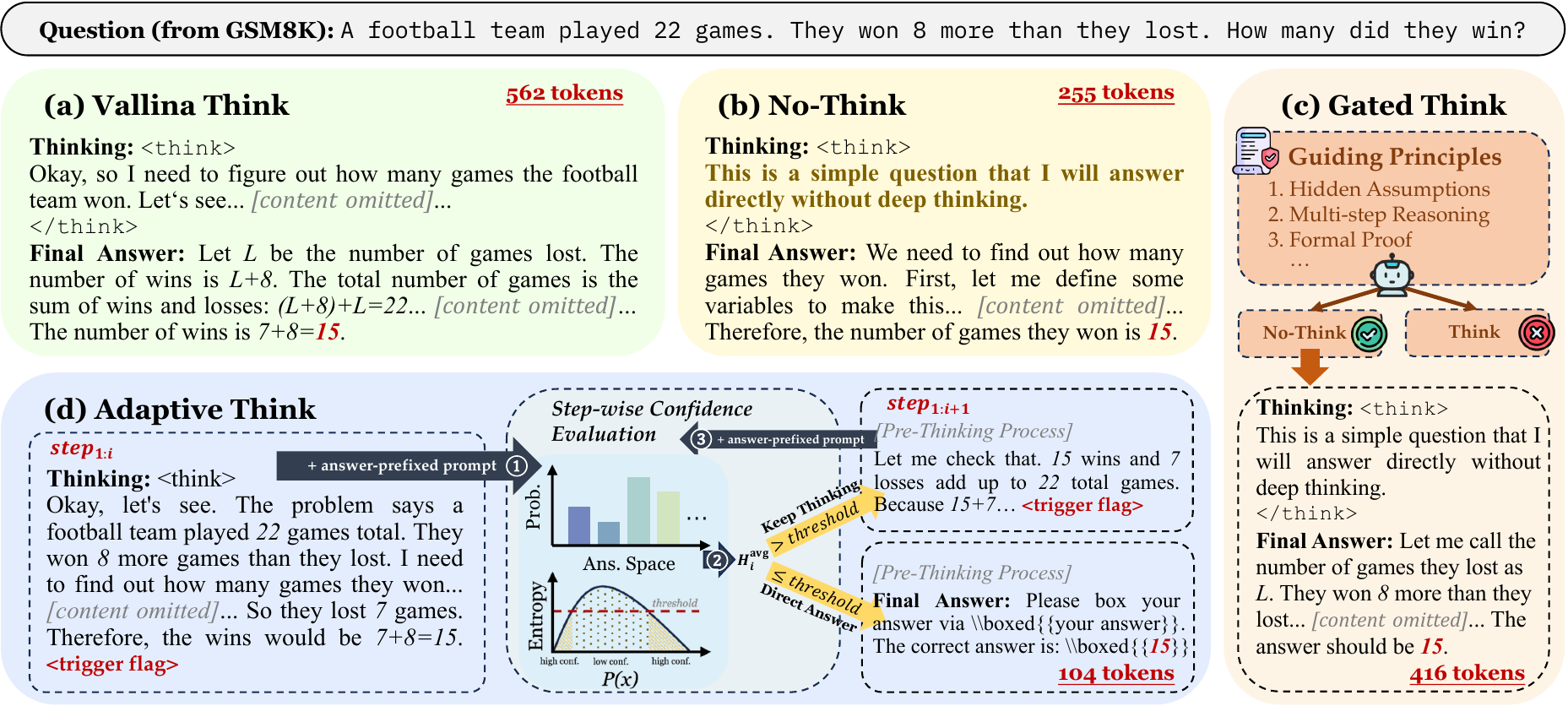}
\caption{
\textbf{An illustration of four thinking modes on a sample question from the GSM8K dataset.}
}
\label{fig:overview}
\end{figure}

Modern LRMs differ fundamentally from earlier non-reasoning models in both training and inference paradigms. Traditional models were typically trained with task-specific supervision to imitate step-by-step reasoning implicitly~\citep{luong2024reft, pang2025bolt}, while modern LRMs are increasingly trained via reinforcement learning to develop general-purpose reasoning capabilities~\citep{guo2025deepseek}. At inference time, these models no longer rely solely on internal heuristics but instead generate explicit reasoning traces, often marked by structured tokens such as \texttt{<think>} and \texttt{</think>}. This shift enables more controllable and interpretable reasoning, opening new avenues for modulating the reasoning process dynamically. Based on this paradigm, we design and evaluate several distinct reasoning modes, as shown in~\autoref{fig:overview}.

\textbf{(a) Vanilla Think.}
It represents the model’s default reasoning pattern, in which it first engages in an extended chain-of-thought process in response to a given question, generating intermediate reasoning steps before eventually producing a final answer based on the full context of its prior thinking.

\textbf{(b) No-Think.}
While most current reasoning models are designed to perform detailed reasoning before producing an answer, it is possible to steer the model toward bypassing this process by modifying the chat template. A common strategy involves forcing the thinking box to remain empty during decoding~\citep{qwen3}. However, we find that using the following prompt more effectively encourages the model to adopt a non-reasoning mode when generating its response.

\begin{lstlisting}[style=promptstyle]
<think>
This is a simple question that I will answer directly without deep thinking.
</think>
\end{lstlisting}

\textbf{(c) Gated Think.}
This setting represents a hybrid of the Vanilla Think and No-Think modes. Given a question, the model is prompted to first assess whether deep thinking is necessary—typically performing this assessment in a no-think mode. To guide this process, we design a heuristic framework that considers several factors, such as whether the question requires inference beyond surface-level cues, involves multi-step reasoning or information synthesis, demands rigorous logical or mathematical justification, presents multiple plausible strategies, or calls for hypothesis-driven analysis. Based on this assessment, the model proceeds in either deep thinking or direct-answer mode. Detailed criteria and prompt are provided in the {\color{citecolor}Appendix C.4}.

\textbf{(d) Adaptive Think.}
Empirical results in~\autoref{sec.3.4} reveal that information bias with respect to the correct reasoning path tends to accumulate as the response length increases. Each reasoning step contributes to reducing entropy over the answer space and increasing confidence in the correct answer, forming clear trends. Since entropy reflects the model's uncertainty over the answer distribution, we propose an Adaptive Think strategy to dynamically decide when to terminate reasoning. After each intermediate reasoning step, the model computes the average entropy $H_i^{\rm avg} = \frac{1}{l}\sum_{i=1}^l H_i$ over the answer space. Reasoning is terminated early once the average entropy falls below a confidence threshold, which is parameterized by a hyperparameter $\alpha \in [0,1]$—with smaller values of $\alpha$ corresponding to stricter entropy thresholds. To formalize this, we note that the entropy of a discrete distribution (i.e., the function $-p \log_2 p$) is upper-bounded by $1/e \ln 2$ when $p \in (0, 1]$. Using this bound, we define the following stopping criterion at the $i$-th step:
\begin{equation}
\begin{cases}
\text{Output the final answer directly} & \text{if } H_i^{\rm avg} \le \alpha \cdot \frac{1}{e \ln 2} \\
\text{Continue reasoning} & \text{otherwise}
\end{cases}.
\end{equation}
When the model is determined to have reached sufficient confidence and no further thinking is needed, we follow the approach introduced by \citet{muennighoff2025s1}, prompting the model to generate a final response by appending an \texttt{</think>} tag—used only when the model is still within the thinking phase—followed by an answer-prefixed prompt to elicit the final output.

\section{Experiments}
\label{sec.exp}
\subsection{Experimental Settings}
\textbf{Models and Datasets.}
We conduct comprehensive experiments using five language models—two reasoning-augmented models (QwQ-32B and DeepSeek-R1-Distill-Qwen-32B) and three standard models (LLaMA3.1-8B-Instruct, Qwen2.5-7B-Instruct, and Phi-4)—across six diverse benchmarks, including two mathematical datasets of varying difficulty and four benchmarks covering distinct types of reasoning tasks, including GSM8K, AIME2025, MMLU-Pro, MuSR, ProntoQA, and CommonsenseQA. Detailed descriptions of each benchmark are provided in the {\color{citecolor}Appendix C.1}.

\textbf{Implementation Details.}
We employ the high-throughput inference engine vLLM~\citep{kwon2023efficient} to support efficient model reasoning. For all methods, each question is evaluated with five independent inference runs, and the results are averaged to ensure robustness. For all datasets, we either adopt the standard prompts provided in their original papers or construct task-specific prompts tailored to our setting. Further implementation details, including dataset-specific prompt templates and answer extraction methods, are described in the {\color{citecolor}Appendix C.5}.

\subsection{Main Results}
\label{sec.results}

We conduct experiments on two math reasoning benchmarks with different difficulty levels: GSM8K (standard) and AIME2025 (challenging), as shown in~\autoref{tab:result1}. First, reasoning models outperform non-reasoning ones significantly. For instance, on AIME2025, QwQ-32B with Vanilla Think achieves 70.67\% accuracy, far surpassing Phi-4’s 20.00\%. However, this comes at the cost of much higher token usage—reasoning models consume on average 3.4× more tokens on GSM8K and 9.5× more on AIME2025 compared to non-reasoning baselines. Additionally, the more challenging AIME2025 benchmark results in substantially higher average token consumption than GSM8K.

\begin{table}[ht]
  \centering
  \caption{
  \textbf{Performance and efficiency comparison on two math reasoning benchmarks.} Models are evaluated based on accuracy (where higher is preferred) and average token count (where lower is preferred) across four different strategies: Vanilla Think, No-Think, Gated Think, and Adaptive Think. The comparison encompasses both non-reasoning and reasoning models, offering a thorough analysis of the trade-offs between reasoning performance and computational efficiency. 
  }
  \resizebox{0.85\linewidth}{!}{
    \begin{tabular}{P{3.5cm} P{2.5cm} C{1.4cm} C{1.4cm} C{1.4cm} C{1.4cm}}
    \toprule
    \multicolumn{1}{c}{\multirow{2}[2]{*}{\textbf{Models}}} & \multicolumn{1}{c}{\multirow{2}[2]{*}{\textbf{Think Mode}}} & \multicolumn{2}{c}{\textbf{GSM8K}} & \multicolumn{2}{c}{\textbf{AIME2025}} \\
          &       & \textbf{Acc $\uparrow$} & \textbf{\#Token $\downarrow$} & \textbf{Acc $\uparrow$} & \textbf{\#Token $\downarrow$} \\
    \midrule
    \rowcolor[rgb]{ .949,  .949,  .949} \multicolumn{6}{c}{\textit{\textbf{Non-Reasoning Models}}} \\
    \multirow{2}[0]{*}{\textbf{Llama-3.1-8B-Instruct}} & \multicolumn{1}{c}{Base} & 82.35  & 281.90  & 0.00  & 1015.37  \\
          & \multicolumn{1}{c}{CoT} & 81.83  & 295.95  & 0.00  & 1201.35  \\
    \multirow{2}[0]{*}{\textbf{Qwen2.5-7B-Instruct}} & \multicolumn{1}{c}{Base} & 90.58  & 314.62  & 8.00  & 802.10  \\
          & \multicolumn{1}{c}{CoT} & 90.55  & 318.55  & 5.33  & 854.14  \\
    \multirow{2}[1]{*}{\textbf{Phi-4}} & \multicolumn{1}{c}{Base} & 94.50  & 314.96  & 13.33  & 1388.26  \\
          & \multicolumn{1}{c}{CoT} & 94.77  & 335.11  & 20.00  & 1536.40  \\
    \midrule
    \rowcolor[rgb]{ .949,  .949,  .949} \multicolumn{6}{c}{\textit{\textbf{Reasoning Models}}} \\
    \multirow{5}[3]{*}{\textbf{DeepSeek-R1-32B}} & Vanilla Think & 95.09  & 718.81  & 56.67  & 9527.68  \\
          & No-Think & 93.18  & 253.48  & 22.67  & 2321.91  \\
          & Gated Think & 94.24  & 296.75  & 53.33  & 7705.33  \\
\cmidrule{2-6}          & \cellcolor[rgb]{ .922,  .953,  .996}Adaptive Think & \textbf{95.98 } & \textbf{356.30 } & \textbf{57.33 } & \textbf{4765.15 } \\
          & \cellcolor[rgb]{ .867,  .922,  .969}$\Delta$ \textit{vs. Vanilla} & \textit{+0.94\%} & \textit{-50.43\%} & \textit{+1.16\%} & \textit{-49.99\%} \\
    \midrule
    \multirow{5}[4]{*}{\textbf{QwQ-32B}} & Vanilla Think & 97.00  & 1132.32  & 70.67  & 14595.73  \\
          & No-Think & 97.00  & 758.22  & 68.00  & 13290.79  \\
          & Gated Think & 96.81  & 824.55  & 70.00  & 13941.63  \\
\cmidrule{2-6}          & \cellcolor[rgb]{ .922,  .953,  .996}Adaptive Think & \textbf{97.73 } & \textbf{379.80 } & \textbf{71.33 } & \textbf{4633.50 } \\
          & \cellcolor[rgb]{ .867,  .922,  .969}$\Delta$ \textit{vs. Vanilla} & \textit{+0.75\%} & \textit{-66.46\%} & \textit{+0.93\%} & \textit{-68.25\%} \\
    \bottomrule
    \end{tabular}}
  \label{tab:result1}%
\end{table}%

Next, we compare three thinking modes: Vanilla Think, No-Think, and Gated Think. Skipping the reasoning step (No-Think) drastically reduces token usage but also degrades accuracy, especially for DeepSeek-R1-32B. Gated Think offers a trade-off between accuracy and efficiency, falling between Vanilla and No-Think in both metrics, indicating modest gains in efficiency at the cost of performance. Finally, our proposed Adaptive Think strategy outperforms all thinking modes. It slightly improves accuracy across both benchmarks while reducing token usage by an average of 58.78\%, effectively pruning redundant reasoning and significantly enhancing efficiency.

\begin{table}[t]
  \centering
  \caption{
  \textbf{Performance and efficiency comparison on four other reasoning benchmarks.}
  }
  \resizebox{\linewidth}{!}{
    \begin{tabular}{ll C{1.1cm} C{1.34cm} C{1.1cm}C{1.34cm}C{1.1cm}C{1.34cm}C{1.1cm}C{1.34cm}}
    \toprule
    \multicolumn{1}{c}{\multirow{2}[2]{*}{\textbf{Models}}} & \multicolumn{1}{c}{\multirow{2}[2]{*}{\textbf{Think Mode}}} & \multicolumn{2}{c}{\textbf{MMLU-Pro}} & \multicolumn{2}{c}{\textbf{MuSR}} & \multicolumn{2}{c}{\textbf{ProntoQA}} & \multicolumn{2}{c}{\textbf{CommonsenseQA}} \\
          &       & \textbf{Acc $\uparrow$} & \textbf{\#Token $\downarrow$} & \textbf{Acc $\uparrow$} & \textbf{\#Token $\downarrow$} & \textbf{Acc $\uparrow$} & \textbf{\#Token $\downarrow$} & \textbf{Acc $\uparrow$} & \textbf{\#Token $\downarrow$} \\
    \midrule
    \rowcolor[rgb]{ .949,  .949,  .949} \multicolumn{10}{c}{\textit{\textbf{Non-Reasoning Models}}} \\
    \multirow{2}[0]{*}{\textbf{Llama-3.1-8B-Instruct}} & \multicolumn{1}{c}{Base} & 48.86  & 202.96  & 37.20  & 154.55  & 85.32  & 333.49  & 74.41  & 52.44  \\
          & \multicolumn{1}{c}{CoT} & 50.57  & 429.68  & 39.26  & 322.62  & 90.72 & 434.58  & 73.82 & 216.46 \\
    \multirow{2}[0]{*}{\textbf{Qwen2.5-7B-Instruct}} & \multicolumn{1}{c}{Base} & 66.57  & 241.80  & 39.60  & 164.54  & 97.84  & 339.09  & 81.31  & 108.17  \\
          & \multicolumn{1}{c}{CoT} & 66.00  & 433.11  & 39.97  & 330.68  & 98.68  & 385.27  & 80.59  & 233.23  \\
    \multirow{2}[1]{*}{\textbf{Phi-4}} & \multicolumn{1}{c}{Base} & 68.00  & 228.29  & 32.62  & 290.15  & 99.44  & 271.44  & 76.82  & 129.33  \\
          & \multicolumn{1}{c}{CoT} & 72.86  & 872.34  & 33.73  & 633.97  & 99.60  & 337.97  & 78.78  & 283.13  \\
    \midrule
    \rowcolor[rgb]{ .949,  .949,  .949} \multicolumn{10}{c}{\textit{\textbf{Reasoning Models}}} \\
    \multirow{8}[3]{*}{\textbf{DeepSeek-R1-32B}} & Vanilla Think & 81.14  & 951.68  & 50.82  & 815.76  & 98.76  & 621.69  & 83.87  & 447.86  \\
          & No-Think & 68.00  & 208.91  & 44.68  & 152.66  & 96.20  & 241.80  & 80.95  & 102.40  \\
          & Gated Think & 77.14  & 245.41  & 46.30 &  306.20 & 98.40 & 433.42 & 80.92 & 105.20 \\
\cmidrule{2-10}          & \cellcolor[rgb]{ .922,  .953,  .996}Adaptive Think &       &       &       &       &       &       &       &  \\
          & \cellcolor[rgb]{ .922,  .953,  .996} - $\alpha=0.1$ & \textbf{79.43 } & \textbf{521.33 } & \textbf{50.03 } & \textbf{568.26 } & \textbf{99.64}  & \textbf{543.79}  & 84.54  & 136.30  \\
          & \cellcolor[rgb]{ .922,  .953,  .996} - $\alpha=0.2$ & 78.57  & 414.25  & 48.84  & 428.13  & 97.28  & 447.66  & \textbf{84.60 } & \textbf{85.16 } \\
          & \cellcolor[rgb]{ .922,  .953,  .996} - $\alpha=0.3$ & 74.57  & 336.22  & 47.25  & 325.77  & 94.72 & 351.30 & 84.28  & 54.73  \\
          & \cellcolor[rgb]{ .867,  .922,  .969}$\Delta$ \textit{vs. Vanilla} & \textit{-2.11\%} & \textit{-45.22\%} & \textit{-1.55\%} & \textit{-30.34\%} & \textit{+0.89\%} & \textit{-12.53\%} & \textit{+0.87\%} & \textit{-80.99\%} \\
    \midrule
    \multirow{8}[4]{*}{\textbf{QwQ-32B}} & Vanilla Think & 76.29  & 1338.95  & 47.12  & 1685.59  & 99.36  & 1167.05  & 85.27  & 606.19  \\
          & No-Think & 76.29  & 612.17  & 42.38  & 634.91  & 98.76  & 697.75  & 85.00  & 179.00  \\
          & Gated Think & 78.57  & 674.57  & 44.31  & 643.70  & 98.80  & 890.42  & 85.09  & 177.43  \\
\cmidrule{2-10}          & \cellcolor[rgb]{ .922,  .953,  .996}Adaptive Think &       &       &       &       &       &       &       &  \\
          & \cellcolor[rgb]{ .922,  .953,  .996} - $\alpha=0.1$ & \textbf{77.14 } & \textbf{629.33 } &   \textbf{47.86} & \textbf{1077.76}  & \textbf{99.96}  & \textbf{882.64}  & \textbf{86.68 } & \textbf{262.90 } \\
          & \cellcolor[rgb]{ .922,  .953,  .996} - $\alpha=0.2$ & 76.86  & 443.69  & 46.11  & 729.11  & 96.68  & 677.54  & 86.52  & 159.81  \\
          & \cellcolor[rgb]{ .922,  .953,  .996} - $\alpha=0.3$ & 74.00  & 317.34  & 44.87  & 500.66  & 94.72 & 532.08 & 85.81  & 88.51  \\
          & \cellcolor[rgb]{ .867,  .922,  .969}$\Delta$ \textit{vs. Vanilla} & \textit{+1.11\%} & \textit{-53.00\%} & \textit{+1.57\%} &  \textit{-36.06\%} & \textit{+0.60\%} & \textit{-24.37\%} & \textit{+1.65\%} & \textit{-56.63\%} \\
    \bottomrule
    \end{tabular}}
  \label{tab:result2}%
\vspace{-1em}
\end{table}%

In addition, \autoref{tab:result2} presents results on four additional benchmarks spanning diverse reasoning types. On all tasks, the entropy-based Adaptive Think consistently outperforms Vanilla Think for QwQ-32B, with an average accuracy gain of 1.23\% and a 42.52\% reduction in token usage. For DeepSeek-R1-32B, Adaptive Think reduces average token consumption by 49.43\%. We observe a slight drop in accuracy on the MMLU-Pro and MuSR datasets compared to Vanilla Think. We hypothesize that this may be due to DeepSeek-R1-32B being a distilled model rather than one trained with reinforcement learning, which could limit its capacity for autonomous reasoning.

Furthermore, on CommonsenseQA, a benchmark focused on shallow, intuition-based reasoning, Adaptive Think achieves substantial improvements in efficiency. Specifically, QwQ-32B with $\alpha = 0.1$ reduces token usage by 56.63\%, while DeepSeek-R1-32B achieves an even larger reduction of 80.99\%, both compared to their respective Vanilla Think baselines. These tasks typically rely on basic elimination strategies and commonsense priors rather than complex, multi-step reasoning, which often leads to redundant or exploratory computational paths. Adaptive Think's entropy-based control halts early once confidence suffices, reducing costs with minimal accuracy impact.

\subsection{In-Depth Analysis}
\label{sec.analysis}

\begin{figure}[htbp]
    \centering
    \begin{minipage}[t]{0.45\textwidth}
        \centering
        \includegraphics[width=\linewidth]{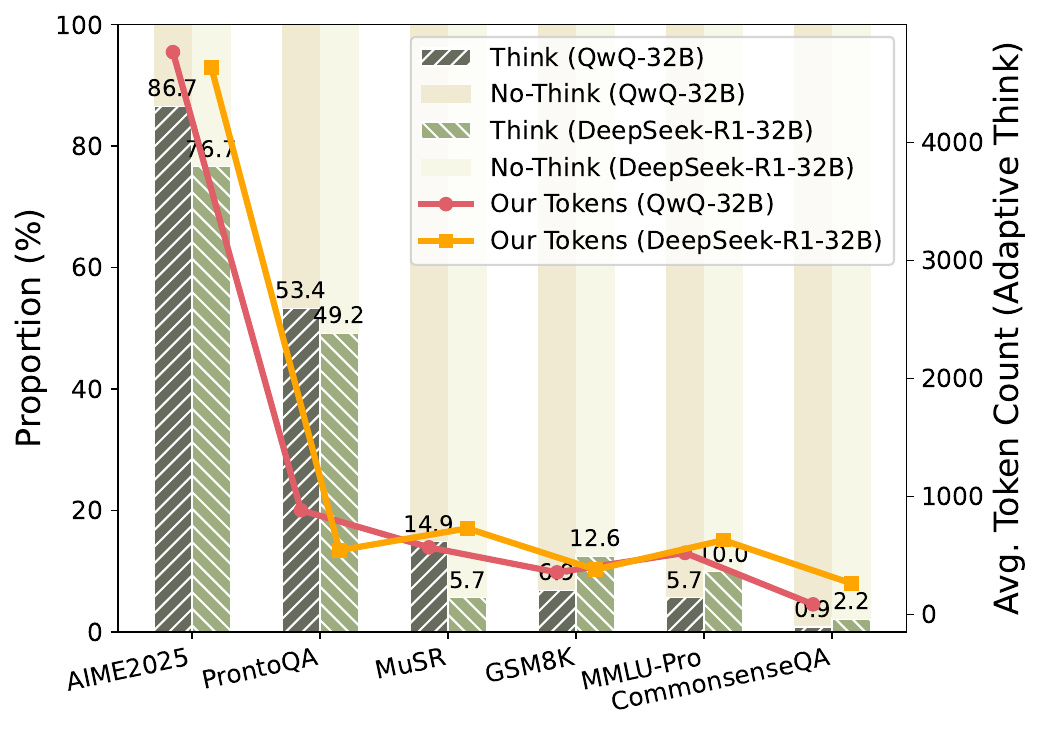}
        \caption{Proportion of think vs. no-think samples in Gate Think mode and corresponding token usage under Adaptive Think.}
        \label{fig:analysis1}
    \end{minipage}
    \hspace{0.035\textwidth}
    \begin{minipage}[t]{0.45\textwidth}
        \centering
        \includegraphics[width=\linewidth]{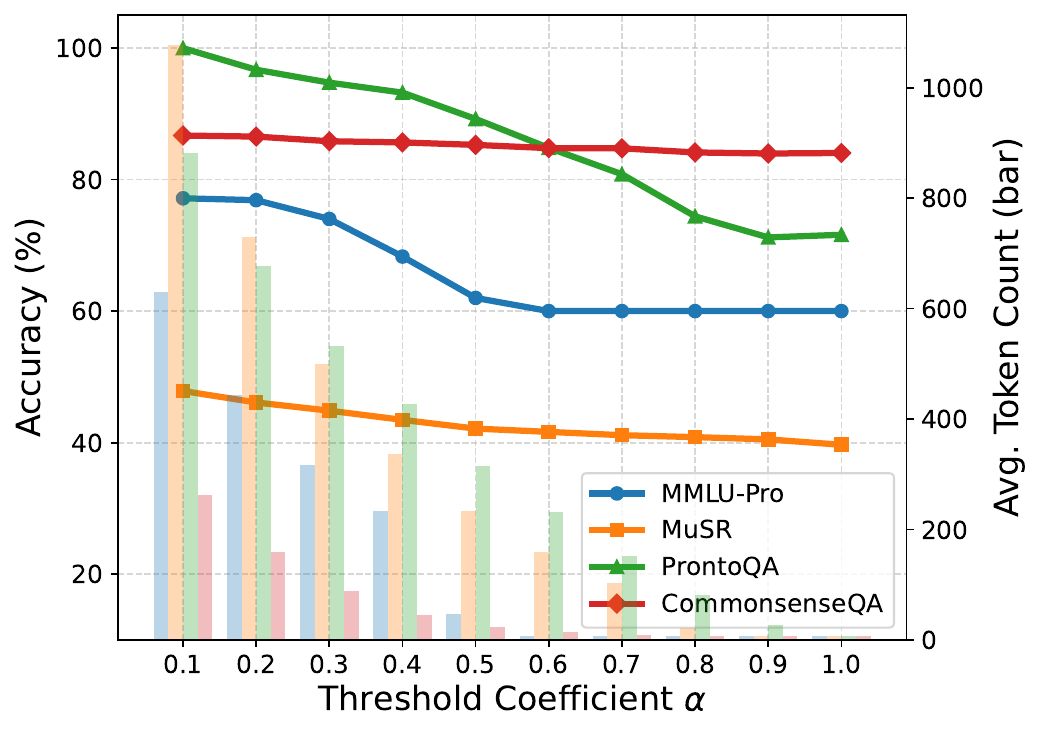}
        \caption{Effect of parameter $\alpha$ on accuracy and token count, showing the trade-off between reasoning performance and efficiency.}
        \label{fig:analysis2}
    \end{minipage}
\end{figure}

\textbf{To Think or Not to Think?}
We analyze the "think" vs. "no-think" decisions under the \textit{Gate Think} setting to assess the model’s ability to adapt reasoning to task difficulty in~\autoref{fig:analysis1}. On AIME2025, which requires strong mathematical reasoning, QwQ-32B and DeepSeek-R1-32B engage in "think" mode for 86.7\% and 76.7\% of samples, respectively. In contrast, for CommonsenseQA—dominated by superficial commonsense cues—these proportions drop to 0.9\% and 2.2\%. This demonstrates the models’ ability to selectively allocate reasoning based on task complexity. Notably, the average token count under Adaptive Think mirrors this pattern, with more computation allocated to harder tasks. This reflects the core strength of \textit{Adaptive Think}: it dynamically adjusts reasoning effort to match problem difficulty, improving efficiency without compromising performance.

\textbf{How Much Thinking is Enough?} 
We further examine how varying the confidence threshold coefficient $\alpha$ in \textit{Adaptive Think} impacts accuracy and token efficiency across tasks (\autoref{fig:analysis2}). Results show that optimal reasoning depth is task-dependent. For logic- and knowledge-intensive benchmarks such as ProntoQA and MMLU-Pro, higher thresholds are critical—premature stopping leads to significant accuracy drops (e.g., from 99.96\% to 71.60\% on ProntoQA and from 77.14\% to 60.00\% on MMLU-Pro). These tasks demand deeper reasoning to resolve ambiguity or recall fine-grained knowledge. In contrast, soft-reasoning tasks such as CommonsenseQA and MuSR exhibit greater robustness to early stopping. Due to their reliance on surface-level cues or redundant contextual information, these tasks allow models to make confident decisions early in the reasoning process. As a result, increasing $\alpha$ leads to minimal accuracy degradation while significantly reducing token consumption, highlighting opportunities for efficiency gains in low-complexity scenarios.

\section{Conclusion}
This paper revisits inefficient reasoning in LRMs through an information-theoretic lens. While extended reasoning chains are often used to improve accuracy, we find that longer outputs often lead to higher bias and semantic redundancy. By introducing InfoBias and InfoGain, we reveal that excessive reasoning often introduces semantic redundancy with limited benefit. Building on these insights, we introduce an entropy-based Adaptive Think strategy that dynamically halts reasoning once confidence is sufficiently high, enabling models to allocate effort based on task complexity while maintaining competitive accuracy. Experiments across diverse tasks and models show that Adaptive Think offers a promising trade-off between efficiency and performance, allowing models to reason selectively—thinking more when necessary, and less when intuition suffices.

\bibliographystyle{plainnat}
\bibliography{references}


\newpage
\appendix
\part{}
\section*{\centering \LARGE{Appendix}}
\mtcsettitle{parttoc}{}
\parttoc

\clearpage

\section{Related Work}
\paragraph{Information-Theoretic Analyses of Language Models}
Information theory provides a principled foundation for analyzing machine learning systems, especially in understanding generalization, uncertainty, and learning dynamics. Classical works apply mutual information and related measures to characterize generalization performance in deep learning~\citep{russo2016controlling,xu2017information}, as well as to clarify the structure of unsupervised learning objectives~\citep{slonim2002unsupervised} and summarization tasks~\citep{west2019bottlesum}.

More recently, these tools have been adapted to LLMs, where entropy-based methods help characterize and diagnose model behavior. For instance, \citet{ton2024understanding} and \citet{gan2025rethinking} use information-theoretic frameworks to study the alignment and reliability of LLM reasoning. A notable development is the introduction of semantic entropy, which quantifies the variability of meaning across possible generations to detect hallucinations and semantic inconsistencies~\citep{farquhar2024detecting}. Semantic Entropy Probes (SEPs) further improve efficiency by estimating this uncertainty directly from intermediate hidden states, without requiring multiple generations~\citep{kossen2024semantic}. Beyond output quality, entropy has also been used to investigate the internal reasoning process of LLMs. \citet{ali2025entropy}, for example, propose entropy-based probes to localize failure points in multi-step reasoning chains, offering a supervision-free alternative to error attribution.

\paragraph{Efficiency and Adaptivity in LLM Reasoning}
As reasoning tasks become more complex and LLMs more capable, managing the cost-performance trade-off has emerged as a critical research theme. Several works propose adaptive reasoning frameworks that dynamically adjust the number of reasoning steps (e.g., chain-of-thought length) based on input difficulty or intermediate confidence~\citep{han2024token,pan2024dynathink,shen2025dast,xu2024adaption}. These approaches often aim to reduce unnecessary computation while preserving answer quality. Complementary to this, early exit strategies enable models to halt generation once sufficient evidence or confidence has been gathered~\citep{yang2025dynamic,damani2024learning,wang2024make}. Other resource-aware designs include complexity-aware token allocation\citep{qu2025optimizing} and routing across model sizes or reasoning depths\citep{kirchner2024prover}, aligning inference cost with task demands.

A parallel line of research revisits the assumption that longer reasoning is always better. Empirical studies show diminishing or even negative returns from overly long CoTs, particularly in high-capacity models prone to spurious logic and hallucinations~\citep{wu2025more,yang2025towards,chen2024not}. This “overthinking” phenomenon suggests that the optimal reasoning depth is both task- and model-dependent~\citep{su2025between}. Interestingly, short or minimal CoTs often achieve comparable performance, especially outside symbolic or mathematical domains~\citep{ballon2025relationship,jin2024impact,sprague2024cot}.

Lastly, some recent perspectives challenge the necessity of explicit chain-of-thought reasoning altogether. Studies show that models can exhibit reasoning-like behavior even without intermediate thought supervision or explicit stepwise prompts~\citep{ma2025reasoning,sui2025stop}. These findings reinforce the need for flexible, confidence-aware mechanisms that can dynamically modulate reasoning depth—potentially without committing to rigid CoT formats.

\section{Limitations and Future Work}
\subsection{Model and Task Selection Constraints}
Adaptive Think requires access to a model’s next‐token probability distribution, so we evaluate it on open‐source deployments that expose this interface. For closed‐source models—such as OpenAI’s o1—we can only employ the sampling‐based approximation method from~\citet{farquhar2024detecting} to estimate the answer‐space distribution, limiting us to analytical assessments of reasoning efficiency. While prior methods focus on multiple‐choice tasks, we broaden Adaptive Think to free‐response benchmarks like GSM8K and AIME2025. Leveraging a tree‐search algorithm to derive answer‐space distributions allows us to rigorously measure entropy reduction through model reasoning. However, for truly open‐ended questions—where no single “correct” answer exists—Adaptive Think cannot yet optimize reasoning efficiency, and, to our knowledge, no existing work has tackled this challenge. We outline this as a key avenue for future investigation.

\subsection{From Output- to Model-Oriented Optimization}
Adaptive Think offers a plug‑and‑play mechanism that dynamically reduces unnecessary reasoning steps and sequence length during inference. This stands in fundamental contrast to model‑based efficient reasoning, which seeks to compress full‑length reasoning models into more concise variants or to train inherently efficient reasoning architectures from scratch. While Adaptive Think provides a lightweight approach to mitigate a model’s tendency toward overthinking, model‑based methods directly enhance the core reasoning capacity and efficiency of the model itself. However, reducing inference cost solely through output manipulation leaves unaddressed the underlying architectural and algorithmic inefficiencies that limit scalability and adaptability across tasks. Transitioning from output‑oriented to model‑oriented optimization is therefore crucial: by redesigning model internals—such as attention mechanisms, intermediate representation formats, and gradient flow pathways—we can achieve more substantial, generalizable gains in reasoning speed, resource usage, and performance consistency. Accordingly, our next phase of work will investigate model‑centric techniques for deeper and more robust improvements in inference efficiency.

\section{Implementation Details}
\subsection{Models and Datasets}
We conduct comprehensive experiments using two reasoning-augmented models:

\begin{itemize}
    \item \textbf{QwQ-32B}~\citep{qwq32b}: a 32B-parameter language model developed by Alibaba's Qwen team, emphasizing advanced reasoning capabilities. It features a 32K token context length and demonstrates performance comparable to OpenAI's o1 model on several benchmarks. The model is designed to embody principles of curiosity and reflection, aiming to enhance analytical reasoning during responses.
    \item \textbf{DeepSeek-R1-Distill-Qwen-32B}~\citep{guo2025deepseek}: a distilled version of the DeepSeek-R1 model, fine-tuned on synthetic data generated by the original R1 model. This variant leverages the Qwen architecture and benefits from reinforcement learning techniques to enhance reasoning capabilities. The distillation process aims to retain the reasoning strengths of DeepSeek-R1 while improving efficiency and accessibility.
\end{itemize}

To evaluate the generality and robustness of our methods, we employ six reasoning-focused benchmarks spanning diverse domains and cognitive requirements. Below, we provide detailed descriptions of each dataset:

\begin{itemize}
    \item \textbf{GSM8K}~\citep{cobbe2021training} (Elementary mathematics)\hfill \\
    \textbf{Reasoning Type:} Multi-step numerical reasoning \\
    \textbf{Description:} GSM8K is a dataset of 8.5K high-quality, linguistically diverse grade school math word problems. Solving each question typically requires several steps of numerical reasoning, often involving intermediate arithmetic operations. The dataset has become a standard testbed for evaluating the chain-of-thought capabilities of language models, especially in structured, algorithmic domains.

    \item \textbf{AIME2025} (Advanced competition mathematics)\hfill \\
    \textbf{Reasoning Type:} Symbolic, multi-step, and abstract reasoning \\
    \textbf{Description:} This benchmark consists of problems from the 2025 American Invitational Mathematics Examination (AIME) I and II, a prestigious U.S. math competition for high school students. Compared to GSM8K, AIME2025 features significantly higher problem complexity, demanding more abstract algebraic manipulation, geometric insight, and symbolic reasoning. It serves as a rigorous test of deep mathematical reasoning.

    \item \textbf{MMLU-Pro}~\citep{wang2024mmlu} (General knowledge and academic reasoning)\hfill \\
    \textbf{Reasoning Type:} Knowledge-intensive multi-hop reasoning \\
    \textbf{Description:} MMLU-Pro is an enhanced version of the original MMLU benchmark~\citep{hendrycks2020measuring}, which contains questions from 57 diverse academic subjects. MMLU-Pro focuses on more complex, multi-hop questions that test both factual knowledge and the ability to integrate information across domains. It is designed to reflect real-world professional exam scenarios, such as medical, legal, or scientific reasoning.

    \item \textbf{MuSR}~\citep{sprague2023musr} (Narrative comprehension)\hfill \\
    \textbf{Reasoning Type:} Soft, multi-step reasoning over long contexts \\
    \textbf{Description:} MuSR (Multi-step Soft Reasoning) features long-form (~1,000 words) natural language narratives, requiring the model to reason about evolving relationships, causality, and world knowledge. Each instance poses a series of questions that depend on the entire story, emphasizing temporal coherence, contextual memory, and soft inference rather than purely symbolic logic.

    \item \textbf{ProntoQA}~\citep{saparov2022language} (Logical deduction)\hfill \\
    \textbf{Reasoning Type:} Symbolic and deductive reasoning \\
    \textbf{Description:} ProntoQA is a synthetic benchmark constructed to systematically evaluate deductive reasoning capabilities in LLMs. The dataset includes logic puzzles framed in natural language, each of which has a unique, deterministically derivable correct answer. It is especially useful for probing consistency, error propagation, and how well models can follow logical implications.

    \item \textbf{CommonsenseQA}~\citep{talmor2018commonsenseqa} (Commonsense knowledge)\hfill \\
    \textbf{Reasoning Type:} Heuristic and intuitive reasoning \\
    \textbf{Description:} CommonsenseQA challenges models to answer questions that require everyday commonsense understanding, typically in the absence of direct textual evidence. It evaluates a model’s ability to use prior knowledge and intuitive judgment to select the most plausible answer among distractors, making it a key benchmark for human-aligned reasoning.
\end{itemize}

\subsection{Information Bias Calculation Settings}
First, we utilize the GSM8K dataset, prompting each model to generate 10 responses per question to obtain a sample-based estimation of the random variable $S$. To ensure consistency in response generation, we employ the following standardized reasoning prompt across all evaluated LLMs:
\begin{lstlisting}[style=promptstyle]
Please answer the question step by step. Remember to box your final answer via $\\boxed{your answer}$. If there is no correct answer, give a random answer.
\end{lstlisting}
Furthermore, to obtain correct reasoning paths $T$ for each question, we leverage the fact that GSM8K provides gold-standard step-by-step solutions. Following the approach in \citet{gan2025rethinking}, we use the Llama3.1-70B-Instruct model to paraphrase each gold solution 10 times. The resulting paraphrases are aggregated to form a sample-based estimation of the random variable $T$. For paraphrasing the ground-truth answers, the following prompt was used:
\begin{lstlisting}[style=promptstyle]
You will be given a problem-solving process. Please rewrite this process without changing its logic or content. Ensure that the output includes only the rewritten process and nothing else.

Problem-Solving Process: {input}

Rewritten Process:
\end{lstlisting}
Subsequently, we estimate the mutual dependence between the random variables $S$ and $T$ using the Hilbert-Schmidt Independence Criterion (HSIC). HSIC provides a non-parametric measure of statistical dependence by projecting the data into a reproducing kernel Hilbert space and quantifying the cross-covariance between the transformed variables. We employ the Gaussian kernel for this purpose, as it offers greater expressiveness for capturing complex, nonlinear relationships compared to linear or inverse multiquadratic (IMQ) kernels. To determine an appropriate bandwidth parameter $\sigma$, we follow a common heuristic based on the median of pairwise Euclidean distances among samples.

To mitigate potential bias introduced by varying response lengths, we normalize the raw HSIC scores by the number of tokens in each response. This normalization yields a per-token dependency measure, allowing for equitable comparison. The overall setup is specifically designed to evaluate the alignment between model-generated reasoning paths and reference derivations.

\subsection{Information Gain Calculation Settings}
\label{app.infogain-settings}
To accurately estimate the conditional probability distribution over the final answer space at a specific stage of reasoning—as well as the corresponding entropy—we concatenate the question \( Q \), the intermediate reasoning steps up to step \( i \), denoted as \( s_{1:i} \), and an answer-prompting suffix. This composite prompt is then used to compute the token-level probabilities of subsequent model outputs. For instance, in the case of multiple-choice questions, we illustrate this procedure using an example from the MMLU-Pro dataset:
\begin{lstlisting}[style=promptstyle]
Question: The symmetric group $S_n$ has $\n\\factorial{n}$ elements, hence it is not true that $S_{10}$ has 10 elements.\nFind the characteristic of the ring 2Z.\nA. 0\nB. 30\nC. 3\nD. 10\nE. 12\nF. 50\nG. 2\nH. 100\nI. 20\nJ. 5

<think>
Okay, so I need to find the characteristic of the ring 2Z and choose the correct option from the given list. Hmm, let me recall what the characteristic of a ring means. The characteristic is the smallest positive integer n such that n times the multiplicative identity (1) equals zero in the ring. But wait, 2Z is the ring of even integers, right? So first, does 2Z have a multiplicative identity? ...[content omitted]...
</think>

The correct option letter is:
\end{lstlisting}

For open-ended question answering tasks such as GSM8K and AIME2025, we adopt a different approach. Given a constructed answer-prompt suffix, we perform a tree-based search over the space of possible continuations. Starting from the prompt, we recursively expand the output tokens until a termination symbol (e.g., a closing brace “\}”) is generated, marking a complete candidate answer path. The probability of each candidate answer is computed as the product of the conditional probabilities of the tokens along its corresponding path. Finally, all candidate sequences are ranked by their total probabilities, and the top-$K$ most probable completions are selected as final answer candidates.
\begin{lstlisting}[style=promptstyle]
Question: Steve and Tim decide to see who can get home from school the fastest. Steve lives further away than Tim, so he is allowed to ride his bike. Steve lives 3 miles from the school and can bike at 440 feet per minute. Tim lives 2 miles away from the school. If Tim can ride his skateboard at 264 feet per minute, how long will the winner be waiting at their house before the loser finishes the race?

<think>
Okay, let's see here. So the problem is about Steve and Tim racing to their homes from school. I need to figure out how long the winner will have to wait for the loser. Hmm, right. Let me parse the details again. First, Steve lives farther away and he bikes at 440 feet per minute. Tim lives closer, only 2 miles away, but skates his skateboard at 264 feet per minute. The question is, who gets home first, and then how much time the winner has to wait? Alright, so I think the key here is to calculate the time each person takes to get home and then find the difference between those times. The person with the shorter time is the winner, and the difference will be how long they have to wait. ...[content omitted]...
</think>

Please box your final answer via \\boxed{{your answer}}. The correct answer is: \\boxed{{
\end{lstlisting}

For the visualization method described in \autoref{sec.infogain-cal}, we first normalize the number of reasoning steps for each question in the Vanilla Think mode to a 0–1 range and plot the line chart of metric changes across all questions. We then fit separate curves for correctly and incorrectly answered questions, along with their 95\% confidence intervals (shaded regions represent the confidence bands). For the No Think mode, we compute the average number of output tokens per benchmark and scale the curve proportionally based on its ratio to the average reasoning length in Vanilla Think mode, ensuring a consistent comparison of the overall reasoning process.

\subsection{Detailed Criteria for Gated Think}
In cognitive and educational psychology, a substantial body of work suggests that the structural features of a question—rather than its superficial difficulty—are key predictors of whether deep human reasoning is likely to be invoked. Specifically, when a problem cannot be answered through direct retrieval or one‐step logic, humans tend to engage in multi‐hop reasoning, hypothesis generation, and information synthesis, as exemplified by contrastive policy learning approaches in symbolic domains~\citep{poesia2021contrastive}. Such cognitively demanding tasks activate deeper, sequential processes involving intermediate inference steps. In formal domains like mathematics and logic, tasks that require proofs—such as induction, contradiction, or recursion—are known to elicit structured chains of reasoning~\citep{besold2021neural}. When a problem presents multiple valid solution paths or requires the evaluation of competing strategies, humans instinctively perform mental simulations of plans, assess outcomes, and prune suboptimal branches—hallmarks of strategic reasoning~\citep{hao-etal-2023-reasoning}. Lastly, tasks that involve multi‐variable relationships, such as scientific hypothesis testing or economic modeling, often require systematic modeling, assumption tracking, and iterative validation. These processes map closely to “System 2” reasoning and have recently been formalized in comprehensive causal reasoning benchmarks—CausalBench, for instance, evaluates LLMs’ ability to identify and reason about cause‐and‐effect structures across diverse domains~\citep{zhou2024causalbench}.

Taken together, these findings suggest that the need for deep reasoning is not determined solely by surface difficulty but by structural complexity—such as requirements for synthesis, recursion, proof, and planning. Based on this understanding, we introduce the following prompt to operationalize this judgment process. The five criteria outlined are derived from well‐established cognitive demands associated with deeper human reasoning.
\definecolor{prompt_color}{RGB}{194, 213, 247}

\begin{tcolorbox}[
  title= Decision Criteria for Triggering Deep Think Mode,
  breakable,   
  fonttitle=\bfseries,
  enhanced,                        
  colback=prompt_color!10,           
  colbacktitle=prompt_color,         
  coltitle=black,                 
  colframe=prompt_color!80!black,    
  coltext=black,                  
  boxrule=0.5pt,
  arc=2mm
]

You are an intelligent reasoning assistant. Upon receiving a question, you must determine whether it requires \textbf{Deep Think Mode}---which involves rigorous, multi-step, and systematic complex reasoning.

\vspace{1em}

\textbf{Evaluation Criteria (At least TWO must be met to trigger Deep Think Mode):}
\begin{enumerate}[label=\textbf{\arabic*.}]
  \item \textbf{Cannot be answered directly based on the question itself}  
  \begin{itemize}
    \item The answer is not immediately apparent from general knowledge, simple reasoning, or single-step calculations.
    \item The question requires combining multiple knowledge points, hidden conditions, or assumptions.
  \end{itemize}
  
  \item \textbf{Multi-step reasoning \& information integration}
  \begin{itemize}
    \item The solution involves sequential logical steps, where each step depends on previous conclusions.
    \item Multiple data sources, conditions, or assumptions must be synthesized to derive the final answer.
  \end{itemize}
  
  \item \textbf{Strict mathematical/logical proof or recursive deduction}
  \begin{itemize}
    \item The problem requires formal proof (e.g., deductive reasoning, axiomatic proofs).
    \item It involves recursive reasoning, mathematical induction, or constructing counterexamples.
  \end{itemize}

  \item \textbf{Non-trivial strategy or non-unique solution}
  \begin{itemize}
    \item The question requires evaluating multiple potential solutions and \textbf{choosing the optimal one}.
    \item There may be multiple valid approaches, requiring deep analysis and comparison.
  \end{itemize}

  \item \textbf{Systematic reasoning \& hypothesis-based deduction}
  \begin{itemize}
    \item The question requires establishing hypotheses and systematically deriving conclusions.
    \item Multiple variables and complex relationships are involved, requiring a rigorous analytical process.
  \end{itemize}
\end{enumerate}

\vspace{1em}

\textbf{Output Format:}
\begin{itemize}
  \item \textbf{``YES'' (Deep Think Mode required)}  
  If the question meets at least 2 criteria, return ``YES'' and briefly explain why.

  \item \textbf{``NO'' (Deep Think Mode not required)}  
  If the question only requires basic or short-step reasoning, return ``NO'' and explain why it can be answered directly.
\end{itemize}

\vspace{1em}
\textbf{Examples}

\textbf{Requires Deep Think}
\begin{itemize}
  \item \textbf{Input}: \textit{``Let A, B, and C be three sets. Prove that $A \cap (B \cup C) = (A \cap B) \cup (A \cap C)$.''}  
  \textbf{Output}: ``YES -- This problem involves set operations and requires a formal mathematical proof with multi-step logical deductions.''

  \item \textbf{Input}: \textit{``If the speed of light is the cosmic limit, but the universe is expanding, is it possible for two regions to be permanently unobservable from each other?''}  
  \textbf{Output}: ``YES -- This question involves relativity, cosmology, and hypothesis-based deduction, requiring systematic reasoning.''

  \item \textbf{Input}: \textit{``On an 8x8 chessboard, if two opposite corners are removed, can it be completely covered by 2x1 dominoes?''}  
  \textbf{Output}: ``YES -- This requires constructing a counterexample, analyzing the board’s parity, and recursive reasoning.''
\end{itemize}

\vspace{0.5em}
\textbf{Does Not Require Deep Think}
\begin{itemize}
  \item \textbf{Input}: \textit{``What is $2^{10}$?''}  
  \textbf{Output}: ``NO -- This is a straightforward computation that can be answered directly.''

  \item \textbf{Input}: \textit{``Tom is 5 years older than Alice. Alice is 10 years old. How old is Tom?''}  
  \textbf{Output}: ``NO -- This is a basic arithmetic problem that does not require complex reasoning.''

  \item \textbf{Input}: \textit{``Why is water heavier than oil?''}  
  \textbf{Output}: ``NO -- This is a factual question about density that can be answered using common knowledge.''
\end{itemize}

\end{tcolorbox}

\subsection{Experimental Setup and Implementation Details}
We employ the high-throughput inference engine vLLM as the execution framework, with generation hyperparameters set to a temperature of 0.8, top-p of 1.0, and a repetition penalty of 1.05. For all evaluated methods, we conduct five independent inference runs per question. The final accuracy and token usage are computed by averaging across these runs.

For our Adaptive Think approach, we segment the model's reasoning trajectory into discrete steps based on paragraph boundaries. Specifically, the occurrence of a double newline (``\textbackslash n\textbackslash n'') is treated as a trigger flag, prompting an entropy-based decision on whether to continue the reasoning process. To mitigate noise and ensure meaningful intermediate content, we enforce a minimum length of 120 characters per reasoning step.

It is worth noting that for the two mathematical benchmarks—GSM8K and AIME2025—we define the answer candidate space using a $top-K=5$. Candidate answer sequences are retrieved via a tree search procedure (detailed in~\autoref{app.infogain-settings}), where token continuations are explored recursively. The maximum tree depth is capped at 10, which is sufficient to accommodate the full range of answer lengths found in both benchmarks.

\section{Additional Experimental Results}
\subsection{InfoBias vs. Reasoning Length}
\begin{figure}[ht]
\centering
\includegraphics[width=\textwidth]{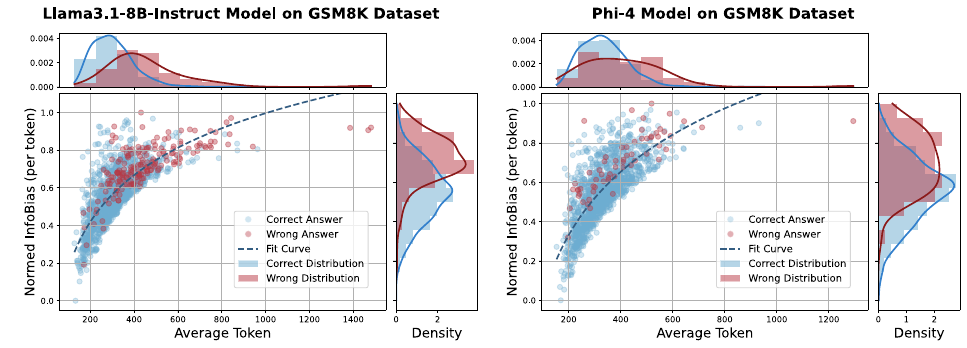}
\caption{
\textbf{Normalized InfoBias per token as a function of average reasoning length for Llama3.1-8B-Instruct and Phi-4 on the GSM8K dataset.} Blue and red points represent instances with correct and incorrect answers, respectively, with density estimates of tokens and InfoBias shown on the top and right. Each subplot illustrates the relationship between reasoning length and InfoBias.
}
\label{fig:infobias-app}
\end{figure}

\autoref{fig:infobias-app} broadens our evaluation by incorporating two additional models—Llama 3.1‑8B‑Instruct and Phi‑4—and uncovers the same systematic pattern identified in~\autoref{fig:infobias}. Specifically, as the length of the reasoning chain increases, the generated outputs progressively diverge from the ground‑truth solution, highlighting a clear trade‑off: each extra token tends to introduce cumulative noise. Quantitatively, we observe a steady rise in solution InfoBias metrics with longer inference trajectories. Crucially, this information drift is not confined to specialized reasoning architectures but also plagues general‑purpose models, underscoring the pervasive challenge of semantic drift in current large‑language systems. These results motivate the need for both output‑level interventions—like adaptive chain‐length control—and deeper, model‑centric optimizations to mitigate drift at its source.

\subsection{InfoGain per Reasoning Step}
\begin{figure}[ht]
\centering
\includegraphics[width=\textwidth]{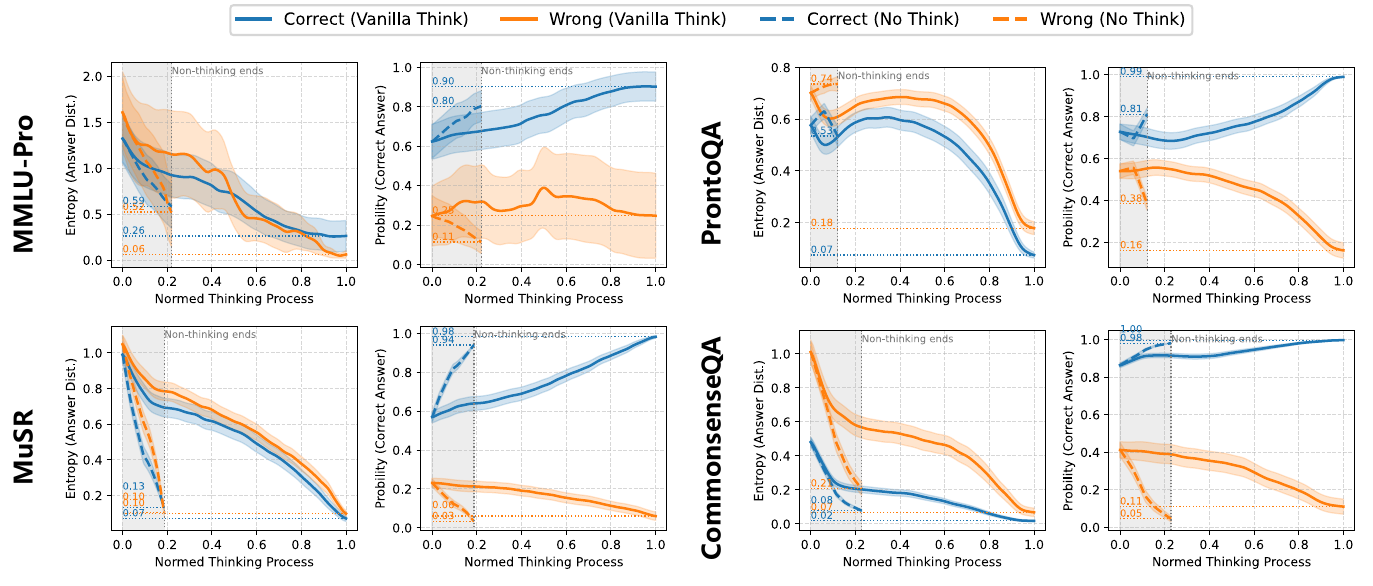}
\caption{
\textbf{Uncertainty dynamics across different reasoning benchmarks for DeepSeek-R1-Distill-32B.} It presents similar thinking dynamics on various benchmarks with QwQ-32B.
}
\label{fig:infogain-app}
\end{figure}

Under the same experimental setup as QwQ‑32B, \autoref{fig:infogain-app} plots how uncertainty‑related metrics evolve throughout the reasoning process of eepSeek‑R1‑Distill‑32B across four distinct types of reasoning tasks.

\section{Case Studies}
In this section, we showcase model reasoning traces from diverse datasets, including case studies where Vanilla Think produced incorrect answers but Adaptive Think, through efficient reasoning optimization, arrives at the correct answers. In the following Vanilla Think reasoning traces, we highlight redundant or irrelevant content—such as repeated or unnecessarily expanded information, background details unrelated to the question, and inconsequential discourse markers—in gray. Critical reasoning steps that lead to incorrect final answers are highlighted in red to indicate misleading or erroneous directions. Since the model already achieves high accuracy on GSM8K and ProntoQA, we selected case studies from the other four benchmarks. Each chosen example was answered incorrectly in all five trials under the Vanilla Think mode but correctly under the Adaptive Think mode.

\subsection{A case from MMLU-Pro}
\definecolor{correct_case_color}{RGB}{200, 235, 190} 
\definecolor{wrong_case_color}{RGB}{255, 225, 180} 

\begin{tcolorbox}[
  title= Case 1: A wrong response from MMLU‑Pro based on the Vanilla Think ,
  breakable,   
  fonttitle=\bfseries,
  enhanced,                        
  colback=wrong_case_color!5,
  colbacktitle=wrong_case_color,
  coltitle=black,
  colframe=wrong_case_color!80!black,
  coltext=black,
  boxrule=0.6pt,
  arc=2mm
]
\textbf{Question:} The study of reality in the broadest sense, an inquiry into the elemental nature of the universe and the things in it, is known as \_\_?\\
A. quantum physics\\
B. metaphysics\\
C. ontology\\
D. phenomenology\\
E. cosmology\\
F. teleology\\
G. epistemology\\
H. axiology\\
I. N/A\\

\textbf{Model Answer:}

\texttt{<think>}

Okay, let's tackle this question. {\color{gray}The user is asking about the study of reality in the broadest sense, specifically an inquiry into the elemental nature of the universe and its contents.} {\color{gray}The options given are from A to J, with some being philosophy branches and others specific sciences.}

First, I need to recall what each term means. {\color{gray}Let me start with the options one by one.}

Quantum physics (A) is a branch of physics dealing with particles at a microscopic level, so probably not the answer here since it's more specific. Metaphysics (B) comes to mind; I remember that metaphysics deals with the fundamental nature of being and the world. Ontology (C) is a part of philosophy that studies being, existence, and reality. So that's very close. Phenomenology (D) is about structures of experience and consciousness, so maybe not. Cosmology (E) is the study of the universe's origin and structure, which is more specific again. Teleology (F) refers to purposes or causes, like the study of design or purpose in nature. Epistemology (G) is about knowledge and justified belief, so that's different. Axiology (H) pertains to values, like ethics and aesthetics. {\color{gray}The rest are N/A.}

The question specifies "the elemental nature of the universe and the things in it." {\color{red}That sounds like ontology, which focuses on being and existence.} But sometimes metaphysics is considered the broader category that includes ontology. {\color{gray}Wait, is ontology a subset of metaphysics? Maybe the answer could be either B or C. Let me think again.} The question says "the study of reality in the broadest sense" and "inquiry into the elemental nature". Metaphysics as a whole covers topics like being, existence, time, causality, etc., while ontology is specifically about being and reality's basic elements. However, sometimes people might refer to metaphysics as the broader field. {\color{gray}Hmm.}

Wait, the exact terms: Metaphysics is often defined as the branch of philosophy investigating the first principles of things, including cosmology, ontology, and teleology. {\color{red}Ontology specifically deals with 'what exists', so if the question is about the elemental nature, ontology might be the direct answer.} However, the phrasing "study of reality in the broadest sense" might align more with metaphysics.

Alternatively, let me check the standard definitions. According to philosophy sources, ontology is the study of being and existence, while metaphysics also includes other areas. {\color{red}Since the question mentions "elemental nature of the universe," maybe ontology is the precise term here.} But sometimes metaphysics is used more broadly. {\color{gray}Let me see examples.} For instance, when philosophers talk about metaphysical questions, they do include things like what is real, the fundamental constituents of the universe, etc. So maybe metaphysics (B) is the right answer because it encompasses ontology within it. 

{\color{red}Alternatively, if the question is pointing directly at the study of existence, then ontology. But I think the common answer expected here would be ontology (C)... So maybe the answer is ontology (C).}

{\color{red}Wait, perhaps the difference is subtle... Therefore, if the question is specifically about the nature of reality's elements, ontology is more precise. But maybe the question considers metaphysics as the broader term.}

{\color{gray}Let me think of how these terms are taught.} In Western philosophy, metaphysics is often called the study of the fundamental nature of reality, being and the world. So perhaps the answer is metaphysics (B). {\color{gray}Wait now I'm confused between B and C.} 

Let me parse the question again: "the study of reality in the broadest sense, an inquiry into the elemental nature of the universe and the things in it" — so that sounds exactly like the definition of metaphysics. For example, when people say metaphysics deals with the fundamental nature of reality, existence, space, time, etc. Ontology is a major branch of metaphysics. So the broader term is metaphysics. Therefore, the correct answer should be B. {\color{red}Hmm, but maybe the question is tricky. Let me think again.}

Wait, the question says "the elemental nature of the universe and the things in it" — elemental in terms of what constitutes them, their basic building blocks. {\color{red}Is that ontology or metaphysics? If it's the elements (like atoms, or substances), that's part of ontology. But metaphysics overall includes those questions. Maybe the best way is to go with the standard textbook answer.}

For example, according to standard definitions, metaphysics is the branch concerned with the fundamental nature of reality. Ontology is a part of metaphysics that deals with existence. So the answer is B, metaphysics. {\color{gray}But I might have mixed this up. Let me try to recall a similar question.} 

{\color{gray}In the GRE or LSAT, such questions often name metaphysics as the study of the fundamental nature of reality. Yes, I think that's the case.} So the correct answer is B. {\color{red}But wait, the options have both B and C. Let me confirm once more. Suppose the question is...}

{\color{red}Alternatively, when someone asks about the fundamental nature of the universe and everything in it, that's typically metaphysics. Since ontology is a subset...}

{\color{red}Maybe the answer expected here is ontology (C). Because ontology is the study of existence and reality's basic elements. Alternatively, metaphysics (B) is the broader term.}

{\color{gray}Let me look for another angle.} The term "ontology" is derived from the Greek "onton" (beings) and "logia" (study). So ontology is the study of beings/existence. Metaphysics is a broader field that includes ontology, cosmology, etc. So if the question is specifically about the study of the elemental nature (the basic elements that compose things), then that's ontology. Hence, the correct answer should be C. {\color{red}So I think I made a mistake earlier thinking it was B. So the answer is C. But I need to be certain.}

{\color{gray}Let me think again. If the question says...} "study of reality in the broadest sense" — metaphysics is broader because it includes not just existence (ontology) but also other issues like causation, possibility. So maybe the answer is B. However, the second part specifies "inquiry into the elemental nature of the universe and the things in it" — that points to the basic constituents and their nature, so ontology. Thus, perhaps the answer is C. {\color{red}I'm a bit torn, but I'll go with ontology (C).}

\texttt{</think>}

After considering the definitions of each term and analyzing the question’s phrasing, the correct answer is ontology (C). The question specifically refers to the "elemental nature of the universe and the things in it," which directly aligns with ontology’s focus on the study of being, existence, and the fundamental categories of reality. While metaphysics (B) is a broader field that includes ontology, the question emphasizes the elemental aspects, making ontology the precise fit. 

$\boxed{C}$\\

\textbf{Correct Answer: B}
\end{tcolorbox}

\begin{tcolorbox}[
  title= Case 1: A correct response from MMLU‑Pro based on the Adaptive Think,
  breakable,   
  fonttitle=\bfseries,
  enhanced,                        
  colback=correct_case_color!5,            
  colbacktitle=correct_case_color,         
  coltitle=black,                  
  colframe=correct_case_color!80!black,    
  coltext=black,                   
  boxrule=0.6pt,
  arc=2mm
]

\textbf{Question:} Same as above.\\

\textbf{Model Answer:}

\texttt{<think>}

Okay, so I need to figure out which term refers to the study of reality in the broadest sense, focusing on the elemental nature of the universe and its contents.

First, I remember that philosophy has several branches. Metaphysics is one of them. I think metaphysics deals with the fundamental nature of being and the world. Ontology is a part of metaphysics, right? Ontology is specifically about what exists and the nature of existence. So maybe the question is pointing towards metaphysics as a broader term?

Then there’s epistemology, which is about knowledge. Phenomenology is the study of structures of consciousness or experience. Cosmology is more about the origin and structure of the universe, often in a scientific context. Teleology deals with purposes or causes. Axiology is about values. Quantum physics is a science, not philosophy. 

The question says "the study of reality in the broadest sense" and "elemental nature of the universe and things in it." That sounds like metaphysics because metaphysics asks questions like "What is existence?" and "What is the fundamental structure of reality?" Ontology is part of that but more specific. Since the question mentions "broadest," maybe B (metaphysics) is better than C (ontology). 

\texttt{</think>}

The correct option letter is: $\boxed{B}$\\

\textbf{Correct Answer: B}

\end{tcolorbox}

\subsection{A case from MuSR}
\definecolor{correct_case_color}{RGB}{200, 235, 190} 
\definecolor{wrong_case_color}{RGB}{255, 225, 180} 

\begin{tcolorbox}[
  title= Case 2: A wrong response from MuSR based on the Vanilla Think ,
  breakable,   
  fonttitle=\bfseries,
  enhanced,                        
  colback=wrong_case_color!5,
  colbacktitle=wrong_case_color,
  coltitle=black,
  colframe=wrong_case_color!80!black,
  coltext=black,
  boxrule=0.6pt,
  arc=2mm
]
\textbf{Question:} In a shocking twist of events, Charlotte is found lifeless in her kitchen, felled by a stun gun, and it’s up to seasoned Detective Winston to unravel the mystery, with prime suspects being Letti and Deborah.

Detective Winston sat at a small, cluttered desk, scanning over the police reports and witness statements regarding the murder scene. His eyes narrowed in on a peculiar piece of information; Charlotte, the victim, had reportedly been stuffed.

"That’s culinary," Winston muttered.

The morning began with a visit to the local pet store, where Letti, a groupie with a penchant for exotic decor, was often spotted buying various animals.

"You’ve seen this woman before?" he asked, flashing his badge and a photograph of Letti towards the shopkeeper.

"Oh, yeah! She comes by often," the shopkeeper said cheerfully. "Always looking for some new pet to tag along with her to those concerts, I guess."

"And what happens to these pets?"

The shopkeeper’s face contorted, his smile fading slightly. "Well... They just disappear, don’t they? It’s odd."

His next stop was the scene of the murder: a secluded kitchen in Charlotte’s home. According to her housekeeper, it was such place which Charlotte and Letti had been seen alone, sharing the early dinner Charlotte had invited her over for.

Suddenly, a lead brought him to a Pawn Shop known for its wide array of self-defense equipment, stun guns in particular. Winston slid the security footage into the player, Letti appearing on the screen, her face a match to the shopkeeper’s description.

She was seen making a purchase, a stun gun catching his trained eye. Making an additional discovery, Winston found a newspaper article, along with some concert footage showing Letti using a stun gun effectively for self-defense during a music event. This was all coming together.

Next, he turned towards a report stating Letti had recently signed up for self-defense classes - the kind that teaches stun gun usage.

His trail returning him to the pet store, Winston had found out that Letti had been at Charlotte’s earlier in the day, according to a tip from a disgruntled neighbor.

Upon hearing about the style in which the stuffing was done, detective Winston noted, Letti seemed to recognize it immediately. Could have been a coincidence, could have not; only time would tell.

Doctor Winston had his work cut out for him, a tangle of clues, and facts left behind, waiting to be pieced together. He could see, however, that the threads of this mystery were starting to knit together. The mystery deepened the more he dug into it, each fact, each clue pulling him further into the complex web spun around the suspect. The only thing left to do was to pull the right string and observe as everything fell into place. With every step, the truth was seemingly closer, yet shrouded with enough obscurity to keep him on his toes.

Winston was met with a distressed and nervous-looking woman when he arrived at Charlotte’s house. Deborah was shaking, a pitiful sight, as she repeated her claim: "I was scared, I was really scared of her." She was a mess, with trembling hands and a trembling voice, but Winston had seen the proof of her fear in the way Charlotte treated her. Harsh treatment by a woman, once her employer and now her victim.

Apparently, Deborah had been scheduled to drive Charlotte that very day; a service she always offered. In all honesty, it seemed Deborah was Charlotte’s only method of transportation. Winston couldn’t help but wonder, though, at the intensity of her fear. Was it because of Charlotte’s abusive behavior, evidenced by bruises and the testimonies of her neighbors? Or was it due to ominous messages that Charlotte had sent Deborah?

Deborah’s fear seemed to have had a history of manifesting physically. The word was out that she had employed a stun gun in the past, for self-protection as Winston understood. Deborah was never far from it, the weapon always at her side.

Just then, Winston noticed a license on her wall. "Used to be a personal security guard, I see," he commented, pointing towards it. Deborah just nodded in agreement, a simple affirmative to Winston’s observation. Strange how life can take unexpected turns, Winston thought.

"Deborah," Winston started, his tone soothing, "can you tell me what exactly happened when Charlotte... died?" Deborah hesitated, her eyes moving towards the stun gun on the table. An eerily similar device was collected from the crime scene.

"Well, I..." she stammered, "I, um, was the last person to see her. I... I found her... in the kitchen." Her voice was barely audible, a whisper barely able to carry the weight of her confession.

Winston didn’t say anything. He simply nodded before standing up, thanking Deborah for her time and promising to be in contact. He left the house, lost in his thoughts. There was much to ponder over in this investigation. After all, the puzzle of the case was only just beginning to reveal itself.\\

Who is the most likely murderer?\\
A. ’Letti’\\
B.  ’Deborah’'\\

\textbf{Model Answer:}

\texttt{<think>}

Okay, let's try to figure this out. So the question is asking who the most likely murderer is between Letti and Deborah based on the given story. I'll need to go through all the details provided and see which suspect has stronger evidence pointing towards them.

First, starting with Letti. The key points here are:

- Letti is a frequent buyer at the pet store, getting exotic animals that "disappear." The mention of Charlotte being "stuffed" might be a clue here. {\color{red}Since Letti deals with animals that disappear, maybe she used a method related to that in the murder.} {\color{gray}But "stuffed" could also refer to the way Charlotte was found, perhaps implying she was literally stuffed like a doll?} {\color{gray}Not sure, but the detective joked it was "culinary," which might suggest something about the method.}

- Letti bought a stun gun from the pawn shop, which matches the murder weapon. There's security footage showing her purchasing it. Also, there's concert footage of her using a stun gun effectively for self-defense. Plus, she enrolled in self-defense classes teaching stun guns. So she has access to the weapon and knows how to use it.

- The neighbor's tip says she was at Charlotte's earlier that day. Since the murder happened at Charlotte's house and the housekeeper mentioned they were seen alone for an early dinner, that places her there. So opportunity is there too.

Now, Deborah's points:

- She was Charlotte's driver and the only one providing transport. She was the last person to see Charlotte alive and found her in the kitchen. However, Deborah was very nervous and claimed she was scared of Charlotte. Testimonies mention Charlotte's abusive behavior towards her, leading to bruises and messages. So possible motive could be fear or retaliation?

- Deborah had a stun gun for self-protection, which was at the scene. The same type was found there. She's a former security guard, so maybe she knows how to use such weapons.

{\color{gray}Wait, the doctor found the stun gun at the scene, but the story mentions both Letti bought one and Deborah carries one. The weapon used was a stun gun, so both have access.} But Letti's possession is documented through the pawn shop purchase, while Deborah's is part of her self-defense.

So Debora's alibi? She says she was the last to see Charlotte and found her. But if she was there, she could have done it. {\color{gray}However, Deborah's motive might stem from being abused, but does that escalate to murder? Maybe.} {\color{red}Letti's motive isn't clear yet. Why would Letti kill Charlotte? Perhaps related to the "stuffing"? Maybe they were involved in something illegal with animals, but Charlotte got in the way? Alternatively, maybe Letti wanted to steal something or there was a personal conflict?}

Other clues: Letti was seen at Charlotte's for a dinner invitation. So they were together alone. That's a perfect opportunity. Deborah was supposed to drive Charlotte that day. {\color{gray}If she drove her there, maybe after the dinner, but Deborah claims to have found Charlotte. Hmm.}

Wait, the housekeeper saw Charlotte and Letti having an early dinner together. So the murder likely happened then. So Letti was there when Charlotte died. Deborah might have come later, or maybe she was present but not mentioned. {\color{gray}Wait, the housekeeper said Charlotte and Letti were alone for the dinner. So Deborah might not have been there at the time of death.} {\color{gray}Unless Deborah had access as the driver. Maybe Deborah was the one who let Letti in or was present, but not sure.}

Deborah's presence at the scene (found her) might mean she arrived afterward. But the murder weapon is a stun gun. Both have access. Letti purchased it, so maybe she used that. {\color{red}The murder weapon could be traced back to the pawn shop purchase.} {\color{gray}Deborah's stun gun is her own, but how would that connect?}

{\color{red}The doctor noted Letti recognized the stuffing method, which might link her to knowing how Charlotte was positioned or something.} {\color{gray}But the story says "stuffing" was culinary, so maybe Letti's expertise in handling animals (like stuffing them?) could be relevant. So Charlotte being "stuffed" could be a play on words, implying Letti did something with preserving or preparing a body, but that's unclear.}

Alternatively, the "stuffing" refers to how Charlotte was positioned, and Letti's familiarity with that suggests she did it. Meanwhile, Deborah's connection is that she found the body but might have been the one who killed her.

So putting it together: Letti has motive {\color{red}(maybe financial, or conflict?)}, opportunity (was there at dinner), means (bought the stun gun). Deborah's motive is abuse, but opportunity might be less direct unless she was nearby. {\color{gray}However, since the housekeeper saw Charlotte and Letti alone during dinner, that's a strong indication that Letti was present when Charlotte was killed.}

Deborah's role as the driver and the last to see her might mean she arrived after the murder and found Charlotte, {\color{gray}but if Deborah had the stun gun, maybe she came afterward and... Wait, but if Letti used the stun gun during the dinner, then let's see timing. Letti and Charlotte are alone. Letti uses the stun gun, kills Charlotte, then leaves. Then Deborah arrives later to drive her, finds her dead. But Deborah's part in that would be coincidental.} {\color{red}Unless Deborah was part of it, but no indication yet.}

Alternatively, Deborah could have acted out of fear, using her stun gun to attack Charlotte, perhaps when she was driving her, {\color{red}but the scene was Charlotte's kitchen. So Deborah might have gone there, arguing, using her stun gun. But why would she be there after the dinner? Maybe she crashed the dinner?} {\color{gray}The housekeeper only mentions Charlotte and Letti being there.}

Another angle: Deborah's possession of a stun gun and being a security guard gives her knowledge. But Letti's recent purchase and training show intent. {\color{gray}The concert footage shows Letti using one effectively, which supports her capability.}

{\color{red}The fact that Deborah was the last to see Charlotte might imply she found the body, but maybe she did it and lied about it. However, Letti has stronger clues: the purchase, presence, and training.} Therefore, the answer might be Letti (option A).

\texttt{</think>}

$\boxed{A}$\\

\textbf{Correct Answer: B}
\end{tcolorbox}

\begin{tcolorbox}[
  title= Case 2: A correct response from MuSR based on the Adaptive Think,
  breakable,   
  fonttitle=\bfseries,
  enhanced,                        
  colback=correct_case_color!5,            
  colbacktitle=correct_case_color,         
  coltitle=black,                  
  colframe=correct_case_color!80!black,    
  coltext=black,                   
  boxrule=0.6pt,
  arc=2mm
]

\textbf{Question:} Same as above.\\

\textbf{Model Answer:}

\texttt{<think>}

Okay, let’s break this down. So Charlotte was found dead in her kitchen, killed by a stun gun. The main suspects are Letti and Deborah. 

First, looking at Letti’s clues: She’s been buying a lot of animals from the pet store, but they disappear. The shopkeeper mentioned that. Then Winston saw security footage of her buying a stun gun at the pawn shop. Also, there’s a newspaper article showing she used a stun gun before at a concert for self-defense. Plus, she signed up for self-defense classes that teach stun gun use. Oh, and the way Charlotte was "stuffed" might be a clue. The detective thought it was culinary, maybe implying Letti’s familiarity with preparing animals (since she buys pets that disappear) led her to stuff Charlotte? That seems a bit odd though. Maybe the "stuffing" refers to how the stun gun was used, like in a specific manner that Letti knows? Now moving on to Deborah. She was Charlotte’s driver and had a history of being abused by Charlotte. Witnesses say Charlotte treated her harshly, even leaving bruises. Deborah has a stun gun for self-defense, which she carries with her. She was the last person to see Charlotte alive and found her body. The murder weapon found at the scene is similar to Deborah’s stun gun. 

Also, Deborah used to be a security guard, so she might know how to use it effectively. Her fear of Charlotte is evident, but why would she kill Charlotte if she was scared? Maybe out of self-defense or retaliation after years of abuse? Wait, but the key points here are the stun gun ownership and opportunity. Both have access, but Letti’s connection is through buying the weapon and knowing how to use it. 

However, Deborah was present at the scene when Charlotte died. The fact that Deborah found the body could mean she was there when it happened, but maybe she did it. Alternatively, she might have come upon the body after someone else did it. Looking deeper, Letti’s motive isn’t clear. Why would she kill Charlotte? The pet store part mentions she buys animals for concerts, but they disappear. 

Maybe Charlotte was involved in that somehow? Like, maybe Charlotte knew something about Letti’s activities, leading Letti to kill her? But the story doesn’t mention any direct link between Charlotte and Letti’s animal purchases beyond them having dinner together. On the other hand, Deborah’s motive is clearer. Charlotte was abusive, maybe Deborah snapped. Plus, she had access as Charlotte’s driver, and had the stun gun. The fact that Deborah was the one who found the body might be a cover-up. If she killed Charlotte, she might have staged it or was present when it happened. The timing matches since she was supposed to drive Charlotte that day. 

Another angle: The "stuffing" comment. The detective thought it was culinary. If Letti works with animals, maybe she used a method similar to stuffing animals, but Charlotte wasn’t an animal. Hmm, perhaps that’s a red herring. Alternatively, maybe the way the stun gun was applied was in a manner Letti learned in her classes, but Deborah, as a former security guard, would also know how. The key evidence points more towards Deborah. She had the means (her stun gun), opportunity (was there), and motive (abuse). Letti’s evidence includes buying the stun gun and prior use, but no clear motive unless there’s something implied about Charlotte’s role in her pet activities. But since Deborah’s situation shows direct conflict and access, she’s more likely. Wait, but wait the article says Letti used a stun gun effectively in the past. And the murder weapon was a stun gun. Letti bought it, so maybe she used it on Charlotte. But Deborah also has one.

\texttt{</think>}

The correct option letter is: $\boxed{B}$\\

\textbf{Correct Answer: B}

\end{tcolorbox}

\subsection{A case from CommonsenseQA}
\definecolor{correct_case_color}{RGB}{200, 235, 190} 
\definecolor{wrong_case_color}{RGB}{255, 225, 180} 

\begin{tcolorbox}[
  title= Case 3: A wrong response from CommonsenseQA based on the Vanilla Think ,
  breakable,   
  fonttitle=\bfseries,
  enhanced,                        
  colback=wrong_case_color!5,
  colbacktitle=wrong_case_color,
  coltitle=black,
  colframe=wrong_case_color!80!black,
  coltext=black,
  boxrule=0.6pt,
  arc=2mm
]
\textbf{Question:} Aside from water and nourishment what does your dog need?\\
A. bone\\
B. charm\\
C. petted\\
D. lots of attention\\
E. walked\\

\textbf{Model Answer:}

\texttt{<think>}

Okay, let's see. The question is asking, aside from water and nourishment, what does a dog need? The options are A. bone, B. charm, C. petted, D. lots of attention, E. walked.

First, I need to think about the basic needs of a dog. The question already mentions water and nourishment, so those are covered. Now, beyond that, dogs require things like exercise, social interaction, affection, maybe veterinary care, a safe environment, etc.

Looking at the options:

A. Bone: While some dogs like bones for chewing, it's not a necessity. Maybe they need something to chew, but bones aren't essential for survival past the basics.

B. Charm: That doesn't make sense. Charm isn't a physical or emotional need for a dog. Maybe "charm" refers to something else, but I can't think of any standard dog need here.

C. Petted: Being petted is a form of affection and comfort. Dogs do benefit from being petted as it provides them with love and reassurance, which is important for their well-being.

D. Lots of attention: Dogs are social animals, so they definitely need attention from their owners. Without enough attention, they might become bored, anxious, or develop behavioral issues.

E. Walked: Exercise is crucial for dogs. Depending on the breed, some need more than others, but regular walks help with physical health and mental stimulation.

{\color{gray} Now, the question says "what does your dog need?" so multiple might seem correct, but we have to choose the best options among the given letters. Let me check the options again. The question might expect one answer. Wait, the options are in singular? The user didn't specify if multiple answers are allowed, but the way the question is phrased, perhaps only one is correct. }

{\color{gray} The options are A to E. The question is from a quiz maybe. So which is more essential besides the basics? Let's think: }

Dogs need exercise (walks), which would cover E. They also need attention and affection (D and C). But between the options, the question might consider 'lots of attention' and 'walked' as both valid, but since it's multiple choice with single answer, which is more critical?

{\color{gray} Alternatively, maybe the intended answer is D. Because "lots of attention" includes petting and interaction, while walking is part of exercise. However, sometimes these questions test knowledge of basic needs like exercise.}

{\color{red} In many such quizzes, the standard answers often include things like exercise, so maybe E is the right answer. }

{\color{gray} Wait another angle: The options are singular, but the question uses "need" which could be plural. But since each option is a single choice, the user expects one correct answer. Let me think again about typical dog needs. Basic needs beyond food and water: }

- Exercise (walking, running)  
- Social interaction (attention, being petted)  
- Healthcare (vaccinations, vet checks, not listed here)  
- Shelter  

So between D and E. The options given are D and E. {\color{gray} The question is probably looking for either. But maybe the most common answer expected here is D or E? }

{\color{red} Hmm, sometimes people might say "lots of attention," but I've heard that dogs need daily walks. Alternatively, perhaps both are correct, but the options are separate. Since the user must pick one, maybe the primary answer expected is E walked because that's a physical necessity, while attention is also important but perhaps considered part of the same category as petting (C). }

{\color{gray} Wait, the options are letters. Let me check each option again. }

Option C is "petted," which is a specific action of stroking, whereas D is "lots of attention" which is broader, including petting, playing, talking to, etc. So maybe D is the better answer.

{\color{red} Alternatively, the question might consider "walked" as essential for health. }

In some contexts, attention is key for their psychological well-being, so without enough attention, a dog can't thrive emotionally. Walking contributes to physical health but might be part of giving attention.

{\color{red} Alternatively, perhaps the answer is E walked, as it's a direct need for exercise, but then D also applies. }

{\color{gray} Hmm. Maybe in the context of the question, the options are presented with one correct answer, and the intended one is D. Or perhaps the original question comes from a source where the answer is D or E. }

Wait, let's think again. The question says "aside from water and nourishment". So necessary besides those. A bone isn't essential. Charm is irrelevant. Between C, D, E.

Petting (C) is a form of affection which is part of attention. So perhaps D encompasses petting, so maybe D is better. But E is also crucial.

{\color{red} But perhaps the best answer is D and E, but since it's multiple-choice with single answer, maybe the question expects D. Or perhaps the answer is E. }

{\color{red} Alternatively, thinking of "needs"—maybe "lots of attention" is more about companionship and mental health, while walks ensure physical activity. Both are needed, but which is more essential? }

{\color{gray} Wait, the options are given, and the user is to choose one. }

{\color{gray} Alternatively, looking up common dog needs: according to standard sources, dogs need exercise (so walks), socialization (attention, interaction), affection, training, etc. So both D and E are correct. But since the options are separate, the correct answer might be D or E. }

{\color{red} Wait the options given are A-E. Do any of the options combine them? No. Then perhaps the question expects one of them. Let me think of possible answer keys. }

{\color{red} Perhaps in this case, the answer is D and E, but since you need to pick one letter, maybe the question's intended answer is D. Or perhaps the most obvious one is E walked. }

{\color{gray} Alternatively, maybe the standard answer in such quizzes is that dogs need lots of attention, so D. But I'm a bit torn. }

{\color{red} Alternatively, considering that 'walked' is a basic physical need, necessary for health, similar to the importance of water and food, so E is the answer. }

{\color{gray} Hmm. I think I'll go with D and E are both correct, but the options may want the primary ones. Given the options, the most likely answers are E and D, but since it's a single choice, maybe the answer is E walked. Or perhaps the question considers "lots of attention" as the correct answer. }

{\color{gray} Wait, looking again, "lots of attention" includes being petted (option C). So maybe D covers both D and C, so D is better. The question is structured so that you have to pick one. }

{\color{red} Alternatively, maybe the answer is E, walked. Since exercise is a basic need. Let me think of the standard list of dog needs. The American Kennel Club or similar sites list exercise as a key component. }

{\color{red} I think the answer is E walked, so boxed E. }

{\color{gray} But I'm not 100\% sure. Alternatively, D might be the intended answer. Since the user asked to choose one, and if I have to pick between D and E, maybe the answer is D. }

{\color{gray} Wait, the question is "what does your dog need?"—so the options are presented as things the dog needs (like a noun). So "attention" is a noun, while "walked" is past tense verb. So maybe the answer is D. }

{\color{gray} Alternatively, walked as in the act of going for walks is a necessity. Maybe the preposition is missing. Like "a walk" would be a noun. But the options are given as walked, petted, etc. }

Given that "lots of attention" is a proper noun phrase, I think D might be the answer intended here.

{\color{gray} Hmm. Alright, I'll go with D. Wait but I'm still conflicted. Let me try to recall other similar questions. Often, in such multiple-choice questions, the options might consider "exercise" (walked) as the correct answer. Because it's a physical need beyond food and water, whereas attention is more about social needs. But both are important. }

{\color{gray} Alternatively, maybe "lots of attention" is a better answer since it's more general. But I'm not entirely certain. }

Another approach: checking the options. If "charm" is B, that's unrelated. "Bone" is not essential, though. "Petted" is a part of attention. So the best choices are D and E. Since both are present, but in the options, they are separate. If I have to pick one, maybe the answer expects D. {\color{gray} Or perhaps the question assumes that "walked" is part of being petted and getting attention. }

{\color{red} Alternatively, maybe the correct answer is E walked because it's a basic biological need (for movement, elimination, etc.) while attention is secondary. }

{\color{gray} Hmm. Maybe I should look for similar questions online. }

{\color{gray} Wait, if I can't access external info, I have to decide based on my knowledge. Since both D and E are correct, but in the options provided, they are separate. The question might be designed to have multiple correct answers, but here you have to choose one. }

{\color{red} Alternatively, perhaps the actual correct answer is D and E, but since that's not an option, maybe the answer is the most essential, which would be walked. }

{\color{red} Alternatively, "lots of attention" is crucial for preventing separation anxiety, so that's essential too. }

{\color{gray} Well, I think I'll go with option E walked. Because maybe the question is from a perspective where exercise is the most overlooked after food/water. }

{\color{gray} Or wait, another thought: The question says "need". Vaccinations and shelter are also needs, but not listed here. The options given are A-E, so among those, D and E are plausible. }

{\color{gray} Maybe "lots of attention" is more encompassing. }

{\color{gray} Alternatively, the answer is D. }

{\color{gray} This is tough, but perhaps the most common answer taught in pet care is that dogs need exercise (walks) and attention. Since the options are separate, maybe the answer is E walked. }

{\color{gray} Wait, perhaps "attention" is included in being petted (C) but D is more comprehensive. So the best two are D and E. }

{\color{red} But given that the user insists on one letter, and given that both are plausible, but I think "walked" is more of a necessary daily requirement. So I'll go with E. }

\texttt{</think>}

$\boxed{E}$\\

\textbf{Correct Answer: D}
\end{tcolorbox}

\begin{tcolorbox}[
  title= Case 3: A correct response from CommonsenseQA based on the Adaptive Think,
  breakable,   
  fonttitle=\bfseries,
  enhanced,                        
  colback=correct_case_color!5,            
  colbacktitle=correct_case_color,         
  coltitle=black,                  
  colframe=correct_case_color!80!black,    
  coltext=black,                   
  boxrule=0.6pt,
  arc=2mm
]

\textbf{Question:} Same as above.\\

\textbf{Model Answer:}

\texttt{<think>}

Okay, let’s see. The question is asking, aside from water and nourishment, what does a dog need? The options are A. bone, B. charm, C. petted, D. lots of attention, E. walked. First, I need to think about the basic needs of a dog. The question already mentions water and nourishment, so those are covered. Now, beyond that, dogs require things like exercise, social interaction, affection, maybe veterinary care, a safe environment, etc. Looking at the options: A. Bone: While some dogs like bones for chewing, it’s not a necessity. Maybe they need something to chew, but bones aren’t essential for survival past the basics. B. Charm: That doesn’t make sense. Charm isn’t a physical or emotional need for a dog. Maybe "charm" refers to something else, but I can’t think of any standard dog need here. C. Petted: Being petted is affection, which is good, but is it a necessity? Probably part of attention. D. Lots of attention: Yes, dogs need social interaction and attention to stay happy and behaved. E. Walked: Exercise is important but not essential for their health.

\texttt{</think>}

The correct option letter is: $\boxed{D}$\\

\textbf{Correct Answer: D}

\end{tcolorbox}

\subsection{A case from AIME2025}
\definecolor{correct_case_color}{RGB}{200, 235, 190} 
\definecolor{wrong_case_color}{RGB}{255, 225, 180} 

\begin{tcolorbox}[
  title= Case 4: A wrong response from AIME2025 based on the Vanilla Think ,
  breakable,   
  fonttitle=\bfseries,
  enhanced,                        
  colback=wrong_case_color!5,
  colbacktitle=wrong_case_color,
  coltitle=black,
  colframe=wrong_case_color!80!black,
  coltext=black,
  boxrule=0.6pt,
  arc=2mm
]
\textbf{Question:} On $\triangle ABC$ points $A,D,E$, and $B$ lie that order on side $\overline{AB}$ with $AD=4, DE=16$, and $EB=8$. Points $A,F,G$, and $C$ lie in that order on side $\overline{AC}$ with $AF=13, FG=52$, and $GC=26$. Let $M$ be the reflection of $D$ through $F$, and let $N$ be the reflection of $G$ through $E$. Quadrilateral $DEGF$ has area 288. Find the area of heptagon $AFNBCEM$.\\

\textbf{Model Answer:}

\texttt{<think>}

Okay, so I have this problem about triangle ABC with points D and E on AB, and F and G on AC. The lengths between these points are given, and there’s some reflection stuff involved. The area of quadrilateral DEGF is 288, and I need to find the area of the heptagon AFNBCEM. Hmm. Let me try to break this down step by step. First, let me sketch the triangle ABC. Points A, D, E, B are on AB in that order. So starting from A, it goes to D (4 units), then D to E is 16 units, and then E to B is 8 units. That means the entire length of AB is AD + DE + EB = 4 + 16 + 8 = 28 units.  Similarly, on AC, the points are A, F, G, C. Starting at A, AF is 13, FG is 52, and GC is 26. So the total length of AC is 13 + 52 + 26 = 91 units.  Now, M is the reflection of D through F. Reflecting a point over another point means that F is the midpoint between D and M. Similarly, N is the reflection of G through E, so E is the midpoint between G and N.  Let me note down coordinates because coordinate geometry might help here. Maybe setting up coordinate axes with point A at the origin? Let me try that. Let’s place point A at (0, 0). Let’s assume AB is along the x-axis, so point B would be at (28, 0). Since AC is another side, let me put point C somewhere in the plane. To make calculations easier, maybe set AC along the y-axis? Wait, but if I do that, then coordinates can be handled more easily. But actually, since both AB and AC are sides of the triangle, unless it’s a right triangle, they might not be perpendicular. Hmm. Maybe better to assign coordinates such that computations can be done using areas. Alternatively, perhaps use barycentric coordinates or area ratios? But the problem mentions reflections and a specific area given for quadrilateral DEGF, which is 288. Let me see if coordinate geometry is feasible here. Let me proceed with coordinates. Let’s set point A as (0,0). Since AB is along the x-axis, point B is at (28,0). Now, point D is 4 units from A, so D is (4,0). Then E is 16 units from D, so moving along AB, E is at (4+16)=20 on the x-axis, so E is (20,0). Then E to B is 8 units, so B is indeed at 28, which matches. For side AC, since points A, F, G, C are along AC. Let me assume AC is along the y-axis. Then point C will be at (0, c) where c is the length of AC, but actually AC has length 91 units? Wait, no. Wait: AF is 13, FG 52, GC 26, so total AC is 91 units. If we set AC along the y-axis, then: Point A is (0,0), F is 13 units along AC, so if AC is vertical, F would be (0,13). But FG is 52 units, so G would be (0,13+52)=(0,65). Then GC is 26 units, so C is (0,65+26)=(0,91). So yes, AC would be along the positive y-axis with C at (0,91). Wait, but then points D is on AB at (4,0) and F is on AC at (0,13). Hmm. Then reflecting D over F: to find M. Reflection over a point: if you reflect D over F, then F is the midpoint between D and M. So coordinates of F are the average of D and M. So coordinates of D are (4,0). Coordinates of F are (0,13). So M must satisfy: ( (4 + $M_x$)/2 , (0 + $M_y$)/2 ) = (0,13) Therefore, (4 + $M_x$)/2 = 0 → 4 + $M_x$ = 0 → $M_x$ = -4 and (0 + $M_y$)/2 = 13 → $M_y$ = 26 Thus, M is at (-4,26). Similarly, N is the reflection of G through E. Point G is on AC at (0,65). Point E is on AB at (20,0). So the reflection of G over E means E is the midpoint between G and N. Coordinates of G: (0,65); coordinates of E: (20,0). So midpoint between G and N is E, so: ( (0 + $N_x$)/2 , (65 + $N_y$)/2 ) = (20,0) Solving: (0 + $N_x$)/2 = 20 → $N_x$ = 40 (65 + $N_y$)/2 = 0 → 65 + $N_y$ = 0 → $N_y$ = -65 Thus, point N is at (40, -65). So now, points M(-4,26), N(40,-65). Need to find the area of heptagon AFNBCEM. Also, need to confirm what are the vertices of heptagon AFNBCEM. Order is A, F, N, B, C, E, M, back to A? Wait, wait heptagon is seven-sided polygon. Let me parse "heptagon AFNBCEM". The vertices in order: A, F, N, B, C, E, M connected in that order, forming a heptagon. Let me make sure the sequence is correct. The name AFNBCEM would imply that starting at A, then F, then N, then B, C, E, M, and back to A. That seems plausible. But before proceeding, let me check if all those points are defined properly. A, F, N: F is on AC, N is the reflection of G over E which is at (40,-65). B is on AB, C is vertex, E is on AB, and M is the reflection over F on the other side. Hmm. So the heptagon includes vertices along AC, jumping out to N, then going to B, then up to C? Maybe, but need to consider their coordinates. But to compute its area, perhaps subtract the areas of the regions not included in the heptagon from the total area of triangle ABC. Alternatively, compute the area directly using coordinates. Since coordinates are established for all points except C is at (0,91), but wait, in my coordinate system, point C is at (0,91). Wait, but if AC is along the y-axis, then point B is at (28,0) and point C is at (0,91). Therefore, triangle ABC is a triangle with coordinates A(0,0), B(28,0), C(0,91). So the area of ABC can be computed as (base * height)/2 = (28 * 91)/2. Let me calculate that later. The problem states that quadrilateral DEGF has area 288. Let me verify that with my coordinate system, perhaps? Because if I can confirm that, then my coordinate setup is okay. Points D(4,0), E(20,0), G(0,65), F(0,13). Wait, DEGF is a quadrilateral. But points D(4,0), E(20,0) are on AB (x-axis), and G(0,65) and F(0,13) are on AC (y-axis). So connecting D to E to G to F and back to D? Wait, connecting DEGF: so the quadrilateral would be D-E-G-F-D? Let me list those coordinates: D(4,0), E(20,0), G(0,65), F(0,13). Connecting these in order. Hmm, so that’s a four-sided figure. Let me compute its area using the coordinates. I can use the shoelace formula for quadrilaterals. Let’s list the coordinates in order: D(4,0), E(20,0), G(0,65), F(0,13), back to D(4,0). Using shoelace formula: Area = 1/2 |$sum_{i=1 to n}$ ($x_i$ $y_{i+1}$ - $x_{i+1}$ $y_i$)| Compute each term: First, D to E: (4,0) to (20,0): Term1: $x_D$ * $y_E$ - $x_E$ * $y_D$ = 4*0 - 20*0 = 0 E to G: (20,0) to (0,65): Term2: 20*65 - 0*0 = 1300 - 0 = 1300 G to F: (0,65) to (0,13): Term3: 0*13 - 0*65 = 0 - 0 = 0 F to D: (0,13) to (4,0): Term4: 0*0 - 4*13 = -52 Sum all terms: 0 + 1300 + 0 -52 = 1248 Take absolute value and divide by 2: |1248| /2 = 624 Wait, but the problem says that quadrilateral DEGF has area 288, but according to my calculation here, it’s 624. That means my coordinate assumption might be wrong. Hmm, so something went wrong here. Probably my coordinate system is incorrect. Wait, perhaps AC isn’t along the y-axis? Maybe I need a different coordinate system because the area of DEGF comes out wrong. Alternative approach: Maybe assigning coordinates differently. Suppose instead that A is still (0,0), B is (28,0), but C is (p, q), so that when I compute DEGF’s area, it gives 288. Then solve for coordinates accordingly. Hmm. That might be more complicated, but necessary. Alternatively, maybe the sides AD=4, DE=16, etc., are along AB, but AC isn’t necessarily aligned with an axis. So perhaps coordinate system isn’t the best way? Wait, let me think again. Maybe if I set coordinates with A at (0,0), B at (28,0), and C at (0, something). But then according to my previous calculation, DEGF area was 624 instead of 288, so that’s inconsistent. Therefore, maybe my assumption of AC being along the y-axis is invalid. Alternatively, maybe both AB and AC are along coordinate axes, but the coordinates need scaling? Alternatively, maybe the coordinates should be set such that DEGF has area 288. Let me try parameterizing. Alternatively, use vectors or mass point? Hmm. Alternatively, use area ratios. Since we have points divided in certain ratios along the sides, perhaps we can express areas in terms of the total area of ABC? Given that DEGF’s area is 288, then maybe the total area of ABC can be found via proportions? Let me try to see. First, note the positions along AB and AC: Along AB: AD=4, DE=16, EB=8. So the division points D divides AB into AD:DB = 4:(24), but since DE is 16, actually the segments are AD=4, DE=16, EB=8. So the ratio from A to D is 4/(4+16+8)=4/28, D to E is 16/28, etc. Similarly along AC: AF=13, FG=52, GC=26. So AF:FG:GC = 13:52:26 = 1:4:2, since 13*1=13, 13*4=52, 13*2=26. So from A to F is 1 part, F to G is 4 parts, G to C is 2 parts. Each part is 13 units. Total length 91, as before. Perhaps coordinate system should be chosen so that the area of DEGF is 288. Let me try again with variables. Let me denote coordinates as follows: Let’s place point A at (0,0), point B at (28,0), and point C at (0,c), making AC along the y-axis. Then as before, points D(4,0), E(20,0), F(0,13), G(0,65). Then quadrilateral DEGF would have coordinates as above. Calculated area was 624, but needs to be 288.  Therefore, discrepancy suggests that my coordinate system is scaled incorrectly. Perhaps the coordinate system I chose assumes that the length along AC is 91 units, but perhaps in reality, the actual coordinates shouldn’t have AC purely along the y-axis. Because otherwise, the area DEGF is too large. Alternatively, maybe the triangle isn’t right-angled, so coordinates can’t have both AB and AC on axes. Hmm. Alternatively, perhaps the coordinates are okay, but I made a mistake in computing the area. Wait let’s recalculate DEGF’s area with my previous coordinates. Coordinates: D: (4, 0) E: (20,0) G: (0,65) F: (0,13) Order of vertices: D-E-G-F-D. Compute shoelace formula: List the coordinates in order and repeat the first at the end: (4,0), (20,0), (0,65), (0,13), (4,0) Calculations: Multiply x by next y, sum them: 4*0 + 20*65 + 0*13 + 0*0 = 0 + 1300 + 0 + 0 = 1300 Then multiply y by next x, sum them: 0*20 + 0*0 + 65*0 + 13*4 = 0 + 0 + 0 + 52 = 52 Subtract the two sums: |1300 - 52| = 1248. Half of that is 624. So same result. Hmm. The problem states it’s 288. So maybe my coordinate system is wrong in assuming right angle at A? Ah, perhaps the triangle isn’t right-angled. So placing C at (0, c) assumes a right angle, but if the triangle isn’t right-angled, then coordinates must be set differently. Hmm, so maybe coordinate system must allow both AB and AC to be in different directions, not necessarily perpendicular. Let me think. Alternative approach: Assign coordinates with A at (0,0), AB along x-axis, so B is (28,0). Let’s let C be at (p, q). Then AC is from (0,0) to (p,q). Then points F, G divide AC into segments AF=13, FG=52, GC=26. So total AC length is 91, so the coordinates of F, G can be determined proportionally. Let me parametrize AC. The total length from A to C is 91, so AF is 13, so F is 1/7 of the way from A to C (since 13/91 = 1/7). Similarly, FG is 52, which is 4/7 of AC, so G is 1/7 + 4/7 = 5/7 from A. Thus coordinates: If point C is (p, q), then: F is ( (1/7)p, (1/7)q ), because moving 1/7 of the way from A to C. G is ( (5/7)p, (5/7)q ). Similarly, points on AB: D is 4 units from A, so since AB is 28 units, D is at (4,0). E is 4 + 16 = 20 from A, so (20,0). Now, quadrilateral DEGF has vertices D(4,0), E(20,0), G( (5p)/7, (5q)/7 ), F( p/7, q/7 ). Need to compute its area and set equal to 288, then maybe solve for p and q, but that seems complex. Also, the area of the heptagon would be expressed in terms of p and q as well. Hmm, but perhaps there is a smarter way. Alternatively, since we’re dealing with areas, perhaps use determinants for areas based on coordinates. The area of quadrilateral DEGF can be calculated using shoelace formula with coordinates in terms of p and q. Let’s try that. Coordinates: D: (4,0) E: (20,0) G: ( (5p/7), (5q/7) ) F: ( (p/7), (q/7) ) Back to D: (4,0) Shoelace formula: Compute sum of $x_i$ $y_{i+1}$ minus $x_{i+1}$ $y_i$. First, write coordinates in order: 1. D: (4, 0) 2. E: (20, 0) 3. G: (5p/7, 5q/7) 4. F: (p/7, q/7) 5. Back to D: (4,0) Calculating the terms: Term1: $x_D$ * $y_E$ - $x_E$ * $y_D$ = 4*0 - 20*0 = 0 Term2: $x_E$ * $y_G$ - $x_G$ * $y_E$ = 20*(5q/7) - (5p/7)*0 = (100q)/7 - 0 = 100q/7 Term3: $x_G$ * $y_F$ - $x_F$ * $y_G$ = (5p/7)*(q/7) - (p/7)*(5q/7) = (5pq)/49 - (5pq)/49 = 0 Term4: $x_F$ * $y_D$ - $x_D$ * $y_F$ = (p/7)*0 - 4*(q/7) = -4q/7 Total sum of terms: 0 + 100q/7 + 0 -4q/7 = (100q -4q)/7 = 96q/7 Absolute value divided by 2: |96q/7| /2 = 48q/7 Set this equal to 288 (the given area of DEGF): 48q/7 = 288 Thus, solving for q: q = (288 *7)/48 = (288/48)*7 = 6*7 = 42 Thus, q=42. Therefore, the y-coordinate of point C is 42. Hmm, interesting. So point C is (p,42).  Now, need to determine p. Since the coordinates of C are (p,42), then AC has length 91 units, which is the distance from A(0,0) to (p,42):  sqrt(p² +42²) = 91 Thus, p² + 1764 = 91² = 8281 Therefore, p² = 8281 -1764 = 6517 Wait, 8281 - 1764: Let me calculate that correctly. 8281 - 1000 = 7281 7281 - 764 = 7281 -700 -64 = 6581 -64= 6517. Yes. Thus, p = sqrt(6517). Hmm. Not a perfect square? Wait 8281 is 91 squared, 42 squared is 1764. Hmm, okay, so p can be either positive or negative, but since it’s a triangle, probably taking positive. However, the actual value of p may not matter for the area calculations, since we might just need the area ratios. Wait, but to find coordinates of other points like M and N, we might need p. Alternatively, perhaps since the area of ABC can be calculated as (base * height)/2. Since AB is the base along x-axis with length 28, the height is the y-coordinate of point C, which is 42. Wait, hold on! Wait, if AB is the base, the height would be the altitude from point C to AB. But in our coordinate system, point C is at (p, 42), so the y-coordinate is 42, so the height from C to AB (which is the x-axis) is indeed 42. Therefore, area of triangle ABC is (28 *42)/2 = 28*21 = 588. Wait, but earlier, with C at (0,91), that was giving area 28*91/2, but here with C at (p,42), the area is 588? Wait, but how does that reconcile with AC length?  Wait, the length of AC is 91, which in coordinate terms is sqrt(p$^2$ + 42$^2$) = 91, so p must satisfy that equation. Let me note that the area of the triangle ABC is 588 regardless of p? Hmm, no, actually the area depends only on the base and height, so if AB is 28 and the height is 42, then yes, the area is fixed at 588. However, the length of AC is constrained by that p, so if the user didn’t specify the triangle is right-angled, but just gives the areas, so this seems okay.  Wait, so perhaps coordinate system is okay, but I need to keep in mind that C has coordinates (sqrt(6517), 42). But sqrt(6517) is approx 80.73... but that might complicate coordinates. Alternatively, perhaps in the problem, the coordinate of C isn’t needed numerically, but since we can work with variables? Alternatively, since we know the area of ABC is 588, perhaps that’s useful. Let me see. We need the area of heptagon AFNBCEM. Maybe instead of calculating the coordinates, we can compute areas of different sections and subtract/add? Alternatively, since the coordinates can be represented in terms of p and q, and we found q=42, so let me see. Wait, perhaps even with the coordinates being (p,42), since the final area might not depend on p? Let me see. Wait, area of the heptagon could potentially be expressed in terms of the triangle’s area minus areas of certain regions. Let me consider the heptagon AFNBCEM. Let me try to visualize the heptagon. Its vertices are A, F, N, B, C, E, M. Starting from A(0,0), going to F( p/7, 42/7 ), since F is 1/7 along AC. Then to N, which is reflection of G through E.  Wait, need coordinates for N and M. Earlier, when I assumed coordinates with C at (0,91), I miscalculated because that led to inconsistency. But now, with C at (p,42), let me recast: Reflecting D over F to get M, and G over E to get N. First, point D is (4,0). Point F is (p/7, 42/7). To find M, the reflection of D over F: The midpoint between D and M is F. Therefore: $F_x$ = ($D_x$ + $M_x$)/2 -> $M_x$ = 2$F_x$ - $D_x$ = 2*(p/7) -4 Similarly, $M_y$ = 2*(42/7) - 0 = 2*(6) = 12 Thus, coordinates of M: (2p/7 -4, 12) Similarly, point G is (5p/7, 30), since G is 5/7 along AC, so coordinates (5p/7, (5/7)*42)= (5p/7, 30). Wait, since q was found to be 42, so 5/7 of 42 is 30. Got it. Point E is (20,0). Reflecting G over E to get N. Midpoint between G and N is E: $E_x$ = ($G_x$ + $N_x$)/2 -> $N_x$ = 2*$E_x$ - $G_x$ = 2*20 - (5p/7) = 40 - 5p/7 Similarly, $N_y$ = 2*$E_y$ - $G_y$ = 2*0 -30 = -30 Thus, coordinates of N: (40 -5p/7, -30 ) Now, having coordinates for all the heptagon’s vertices. So vertices are: A: (0,0) F: (p/7, 6) [since 42/7=6] Wait, 42/7 is 6. So F is (p/7, 6) N: (40 -5p/7, -30) B: (28,0) C: (p,42) E: (20,0) M: (2p/7 -4,12) Back to A: (0,0) To compute the area of heptagon AFNBCEM, I can use the shoelace formula with these coordinates. This might be tedious but manageable. Let me list all coordinates in order: 1. A: (0, 0) 2. F: (p/7, 6) 3. N: (40 -5p/7, -30) 4. B: (28, 0) 5. C: (p, 42) 6. E: (20, 0) 7. M: (2p/7 -4, 12) 8. Back to A: (0,0) Applying shoelace formula: Area = 1/2 |$sum_{i=1 to 7}$ ($x_i$ $y_{i+1}$ - $x_{i+1}$ $y_i$)| Compute each term step by step. Let me denote each vertex as: Vertex 1: A (0,0) Vertex 2: F (p/7,6) Vertex 3: N (40 -5p/7, -30) Vertex4: B(28,0) Vertex5:C (p,42) Vertex6:E (20,0) Vertex7:M (2p/7 -4,12) Vertex8:A (0,0) Now compute terms: Term1: x1*y2 - x2*y1 = 0*6 - (p/7)*0 = 0 -0=0 Term2:x2*y3 -x3*y2 = (p/7)*(-30) - (40 -5p/7)*6 = -30p/7 - [240 - (30p)/7 ] = -30p/7 -240 + 30p/7  = -240 Term3:x3*y4 -x4*y3 = (40 -5p/7)*0 -28*(-30) = 0 + 840 =840 Term4:x4*y5 -x5*y4 =28*42 - p*0=1176 -0=1176 Term5:x5*y6 -x6*y5= p*0 -20*42 =0 -840=-840 Term6:x6*y7 -x7*y6=20*12 - (2p/7 -4)*0=240 -0=240 Term7:x7*y8 -x8*y7= (2p/7 -4)*0 -0*12=0 -0=0 Now, adding all terms: Term1 + Term2 + Term3 + Term4 + Term5 + Term6 + Term7 =0 + (-240) +840 +1176 + (-840)+240 +0 Let me compute step by step: Start with 0 -240 = -240 -240 +840=600 600 + 1176=1776 1776 -840=936 936 +240=1176 Total sum is 1176 Take absolute value (which is still 1176) and divide by 2: Area = 1176 /2 = 588 Wait, but that can’t be. The area of the heptagon is equal to the entire triangle’s area? Because the triangle’s area was 588. That would mean the heptagon covers the whole triangle, but clearly, the heptagon is missing some regions, like quadrilateral DEGF which has area 288. Therefore, this suggests that my calculation is wrong. Hmm, perhaps I messed up the order of the vertices? Let me check again. Wait, perhaps the order of the vertices is incorrect, leading to overlapping areas or incorrect traversal direction. Let me re-express the coordinates with variables, but let me see: Heptagon vertices: A, F, N, B, C, E, M. Connecting A to F is straightforward. From F to N: N is at (40 -5p/7, -30), so that’s below the x-axis. Then from N to B(28,0): moving upwards. Then B to C is going to the point C(p,42). Then C to E is going down to E(20,0). Then E to M: M is at (something,12). So E(20,0) to M(2p/7 -4,12). Then back to A. Hmm, maybe the path crosses over itself? Alternatively, perhaps the shoelace formula got confused because of the negative coordinates. For instance, point N is at (40 -5p/7, -30), which may be at a negative y-value, so when computing the area, the path might loop around in a way that overlaps or subtracts areas. Alternatively, I might have made an arithmetic error in computing the shoelace terms. Let me recalculate each term carefully. Let me redo the terms one by one with clear steps: Vertices: 1. A (0,0) 2. F (p/7, 6) 3. N (40 - (5p)/7, -30) 4. B (28, 0) 5. C (p,42) 6. E (20,0) 7. M ( (2p/7 -4), 12 ) 8. A (0,0) Compute terms for shoelace: Term1: (x1*y2 - x2*y1) x1=0, y2=6; x2=p/7, y1=0 0*6 - (p/7)*0 = 0 -0 =0 Term2: (x2*y3 - x3*y2) x2=p/7, y3= -30; x3=(40 -5p/7), y2=6 Term2: (p/7)(-30) - (40 -5p/7)(6) Calculate each part: First part: (p/7)(-30) = -30p/7 Second part: (40 -5p/7)*6 = 6*40 - 6*(5p/7) = 240 - 30p/7 Thus Term2 = (-30p/7) - (240 -30p/7) = -30p/7 -240 +30p/7 = -240 Same as before. Term3: (x3*y4 - x4*y3) x3=(40 -5p/7), y4=0; x4=28, y3= -30 Term3: (40 -5p/7)*0 -28*(-30) = 0 + 840 = +840 Term4: (x4*y5 -x5*y4) x4=28, y5=42; x5=p, y4=0 28*42 - p*0 = 1176 -0 = +1176 Term5: (x5*y6 -x6*y5) x5=p, y6=0; x6=20, y5=42 p*0 -20*42 =0 -840 = -840 Term6: (x6*y7 -x7*y6) x6=20, y7=12; x7=(2p/7 -4), y6=0 20*12 - (2p/7 -4)*0 = 240 -0= +240 Term7: (x7*y8 -x8*y7) x7=(2p/7 -4), y8=0; x8=0, y7=12 (2p/7 -4)*0 -0*12 = 0-0=0 Adding all terms: 0 -240 +840 +1176 -840 +240 +0 = Let me add sequentially: Start at 0. -240 → total: -240 +840 → 600 +1176 → 1776 -840 → 936 +240 → 1176 Yes, total sum 1176. Half of that is 588. Which is exactly the area of the whole triangle, so that can’t be. So something is wrong here. It implies the heptagon covers the entire triangle, but the problem states that quadrilateral DEGF is inside the triangle with area 288, so the heptagon’s area should be less than 588. Therefore my error must be in the ordering or selection of vertices. Looking at the vertices of the heptagon: AFNBCEM. The points go A to F (on AC), then to N (outside the triangle?), then to B, then to C, to E, then to M (somewhere else), then back to A. Depending on where N and M are, it might be enclosing the entire triangle? Wait, point N is at (40 -5p/7, -30), which has a negative y-coordinate, so outside the triangle below AB. Similarly, point M is at (2p/7 -4,12). Depending on p’s value, its x-coordinate could vary. Since p is sqrt(6517)= approx 80. So let me approximate p. Wait let’s compute p: sqrt(6517): Well 80$^2$ is 6400, so sqrt(6517) =80.73. So p=80.73. Therefore: Compute coordinates for N: $N_x$ =40 - (5p)/7 =40 - (5*80.73)/7 =40 - (403.65)/7=40 -57.66=-17.66 N is at (-17.66, -30). That’s way outside the triangle. Point M: $M_x$ = (2p/7 -4) =(2*80.73/7 -4)=(161.46/7 -4)=23.06 -4=19.06 $M_y$ is 12. So M is around (19,12), which is inside the triangle? Since C is at ~80.73,42, so halfway would be higher, so (19,12) is on the left side. So the heptagon goes from A(0,0) to F (around 11.5,6) since p/7=11.5, then to N(-17.66,-30), then to B(28,0), then to C(approx (80.73,42)), then to E(20,0), then to M(19,12), back to A. This path likely encloses most of the triangle but also includes some areas outside, hence why the shoelace gives the full area. That suggests the order of the vertices might be incorrect, causing the path to traverse outside and then back, leading to overlapping areas canceling out. Hmm. Maybe the order is not correct. The problem says "heptagon AFNBCEM". Let me confirm the order: A first, then F, N, B, C, E, M, back to A. If plotted, that path goes from A to F (up along AC), then to N (way below and left), then jumps to B, then to C, then down to E, then to M (near E), back to A. Such a path would create a star-like shape with overlapping regions. Hence, shoelace might not work unless the polygon is simple (non-intersecting). Therefore, perhaps the vertices are supposed to be ordered differently? Maybe there is a typo or misordering? Alternatively, the problem might require another approach. Alternatively, instead of coordinates, maybe use mass point or area ratios. Wait, given that DEGF has area 288, which we used to find q=42, leading to total area 588. Therefore the heptagon’s area is total area minus the area of some regions. The problem asks for the area of heptagon AFNBCEM. Let me try to figure out which regions are excluded. Looking at the heptagon’s vertices: AFNBCEM. The heptagon excludes the following regions: - The quadrilateral DEGF itself (area 288) Additionally, perhaps other regions? Wait, let me see. Let’s think of the heptagon vertices: - Starting at A, goes to F (on AC), then to N (which is reflection of G over E, outside the triangle), but wait, if N is outside, then the area might include regions outside the triangle. Hmm, this complicates things. Alternatively, perhaps the heptagon is within the triangle? If N is outside, but maybe the problem counts area algebraically? Alternatively, maybe my coordinate approach is not the best. Let me think of another method using reflections and area relations. First, we know the area of the triangle is 588 (from previous calculations with q=42). Therefore, if I can compute the areas of the regions not included in the heptagon and subtract them from 588, that would give the heptagon’s area. What regions are not included in heptagon AFNBCEM? Looking at the vertices: - The heptagon includes points A, F, N, B, C, E, M. So, areas excluded might be: 1. The quadrilateral DEGF (given as 288) 2. The triangle(s) created by points outside the heptagon, such as near M and N? Wait, but N is reflected over E from G, so maybe triangle ENG? Wait, maybe decomposing the figure into known areas. Alternatively, since M is the reflection of D over F, then the segment FM = FD, similarly EN = EG because N is reflection over E. Wait, let me consider the reflections: Reflecting D over F gives M. Since F is on AC, and D is on AB, M is a point such that F is the midpoint between D and M. Similarly, E is the midpoint between G and N. Therefore, perhaps triangles like DFM and GEN are congruent to other triangles? Alternatively, since the problem involves reflections, maybe the areas contributed by M and N can be paired with original areas? Alternatively, considering that the heptagon AFNBCEM consists of several parts: the area from A to F to N to B to C to E to M and back, perhaps breaking into parts: - The polygon AFBC is a quadrilateral. Wait, no, it’s a heptagon with seven vertices. Maybe splitting into simpler shapes? Alternatively, using coordinates but fixing p. Recall that in the coordinate system, we have q=42, p is sqrt(p² +42²)=91, so p=sqrt(91²-42²). Let me compute that exactly: p² = 91² -42² = (91-42)(91+42)= (49)(133)= 49*133. 49*133: 49*(100+30+3)= 4900 + 1470 + 147= 4900+1470=6370+147=6517. So p= sqrt6517. However, perhaps in calculations, since p is a coordinate, but when I plug into shoelace formula, many terms canceled and the total came out independent of p? That suggests maybe the area is indeed 588? Which contradicts the given DEGF area. Alternatively, I might have misunderstood which points form the heptagon. Wait, heptagon AFNBCEM: points A, F, N, B, C, E, M. Wait, if I consider the path A-F-N-B-C-E-M-A, perhaps the area computed via shoelace is actually encompassing the entire triangle plus some external areas, hence why it’s 588. But since the heptagon includes regions outside the triangle (due to point N), perhaps the actual area is different. But the problem doesn’t specify whether the heptagon is entirely within the triangle. Hmm, problem statement says "heptagon AFNBCEM" without qualification, but maybe it’s formed by connecting those points within the triangle? Though N and M might be outside. Wait, the problem states points are on the sides AB and AC, but M is the reflection of D through F, which is on AC, so if F is between A and C, reflecting D would place M outside the triangle on the opposite side of F from D. Similarly, N is reflection over E, so since G is closer to C than to A, reflecting over E (which is on AB) might place N outside the triangle. Therefore the heptagon likely includes areas inside and outside the main triangle. But the problem mentions the area of the heptagon, which could be a net area. Alternatively, perhaps reflections lead to symmetries where areas can be related. Alternatively, the key is to realize that the area of the heptagon can be expressed as the area of the triangle minus the area of quadrilateral DEGF plus areas introduced by the reflections? Not sure. Wait, alternate idea: since M is the reflection of D through F, the area involving M could mirror some area around F. Similarly for N and G over E. Specifically, the reflection of D over F implies that area of triangle FDM equals area of triangle FFD? Wait no, reflection might preserve areas. Let me think: Reflecting point D over point F gives M, so triangle DFE is congruent to triangle MFE? Actually, since F is the midpoint between D and M, line FM = FD, so for any point, say, E, the triangles DFE and MFE would have the same area since they share FE and bases DF and FM which are equal. Similarly for other points. Perhaps the heptagon can be partitioned into areas related to original triangle minus DEGF plus some mirrored areas? Alternatively, notice that: The heptagon AFNBCEM can be seen as the union of the following regions: - Triangle AFM - Quadrilateral FNBC - Triangle BCE - Trapezoid or something? Alternatively, this decomposition might not be straightforward. Alternatively, perhaps the total area of the heptagon can be computed as area of ABC minus area of DEGF plus areas related to the reflections. Wait, the reflections create points M and N outside the triangle. The areas associated with those points may need to be subtracted or added depending on orientation. Alternatively, let’s consider that the problem gives the area of DEGF as 288, which we used to find q=42, leading to area of ABC as 588. If that’s the case, and the heptagon’s area can be derived from the total area minus areas of regions outside the heptagon or something else. But since calculating via coordinates gave the entire area, maybe the problem expects us to realize that the area of the heptagon is the total area minus something? Wait the area of DEGF is 288. If the heptagon excludes DEGF, then 588 - 288 = 300, but that might not account for overlapping regions. Wait but when I did the shoelace formula, I ended up with the entire triangle’s area, implying the heptagon somehow includes everything, which conflicts with the existence of DEGF. Alternatively, perhaps my coordinate system is flawed because of assuming the reflection points extend the polygon to cover the entire triangle? Hmm. Alternatively, maybe the problem is designed so that the area of the heptagon is Area of ABC minus DEGF plus areas of the mirrored parts. Wait, area of the reflections? Reflecting D over F to get M creates a symmetric point. The area of triangle FMD is zero since it’s a straight line? No, M is a reflection, so F is midpoint. But the figure would create areas. Alternatively, perhaps the reflections’ contributions cancel out in the area calculation. Alternatively, let me compute vectorially. Wait, since the problem uses "reflection through a point", which is equivalent to saying that the reflected point is such that the midpoint is the center point. So: Vector equations: Position of M: M = 2F - D Position of N: N = 2E - G Therefore, we can express coordinates of M and N in terms of D,E,F,G. Given that, perhaps I can compute areas using coordinates without needing exact values of p. Wait in my previous shoelace calculation for the heptagon, all terms involving p canceled out, resulting in area = 588 regardless of p, which seems conflicting. But when I set up the coordinates, the area of DEGF gave us q=42, so the triangle area is fixed at 588. The heptagon’s area came out to be the entire triangle’s area. There must be a miscalculation or misunderstanding in vertex order. Alternative approach: Let me try with the coordinate system where AB is along x-axis from (0,0) to (28,0), and point C is at (0,42), making AC along the y-axis. Wait, previously I thought that led to DEGF area 624, but now if C is at (0,42), let’s recompute DEGF’s area in this coordinate system. Wait, if AC is along the y-axis, then coordinates: A(0,0), B(28,0), C(0,42). Then AC length is 42? Wait no, AC length should be 91. Wait, conflict. Wait in this case, if q=42, then AC length is sqrt(p² +42²)=91. Therefore, point C cannot be at (0,42) unless p= sqrt(91²-42²)=sqrt(6517). So with that, I can’t have AC along the y-axis unless I set p to zero, which contradicts. Wait, but if I set p=0, then point C would be (0,42), making AC length 42, but the problem requires AC length 91. So that won’t work. Hmm. So to have AC length 91 with q=42, p must be sqrt(91²-42²). Therefore, in this coordinate system, C is at (sqrt(6517),42). But the problem is that calculations become messy. However, in the shoelace computation earlier for the heptagon, all terms with p canceled out, giving an area independent of p. That is strange. Wait let me see again in the term calculations: Term2 had -240, which was constant. Terms involving p canceled out? Looking at Term2: Term2: (p/7)(-30) - (40 -5p/7)(6) = -30p/7 - [240 -30p/7] = -240. So p cancels. Term3,4,5, etc., had terms where p appeared but eventually canceled. Indeed, Term4 had 28*42 (doesn’t involve p), Term5 had p*0 -20*42, so -840, Term6 had 20*12, etc. All terms involving p canceled except in the terms for M’s coordinates? Wait $M_x$ is (2p/7 -4), so in Term6, x7 is (2p/7 -4), but Term6 was 20*12 - (terms involving M’s coordinates in Term7? Wait Term7 involves x7*y8 which is (2p/7 -4)*0 and y7*x8 which is 12*0, so Term7 is zero. Therefore, the entire computation of the terms shows that all references to p canceled out, resulting in total 1176, halved to 588, so regardless of p, the area is total triangle area. That must mean the polygon loops around covering the entire triangle in such a way that the shoelace formula computes the area as total. Thus, the mistake must be in the vertex ordering for the heptagon causing the path to traverse around the triangle and close back through negative space. Perhaps the correct order requires a different sequence. Wait the problem specifies the heptagon AFNBCEM. Let me think: Start at A, go to F (along AC), then to N. If N is below the triangle, then the edge F to N goes downward. Then to B, which is on the baseline, then to C, up to C, then down to E, then to M, which might be near the middle, then back to A. Alternatively, perhaps the correct area is indeed total area minus DEGF’s area, so 588-288=300, but why? Wait the problem says "quadrilateral DEGF has area 288". If the heptagon doesn’t include DEGF, then subtract it. But according to the vertex list AFNBCEM, DEGF is a separate region not part of the heptagon. So heptagon area = 588 - 288 = 300? Is that possible? Alternatively, maybe there are other regions excluded besides DEGF. For instance, the areas near M and N. Wait, reflecting D over F gives point M beyond F on the opposite side of F from D. Since D is on AB and F is on AC, the point M is located such that it’s mirrored over F. Similarly for N. So the heptagon may exclude regions like DEM and others. Alternatively, maybe the area of the heptagon is (Area of ABC) plus areas of the reflected regions but this is getting too vague. Alternatively, since reflecting points over Midpoints, the areas associated with those reflections can be linked via symmetry. Wait, let me think of the heptagon as follows: - Start at A and go to F, which is 1/7 along AC. - Then to N, which is the reflection of G over E. G is 5/7 along AC at (5p/7,30). Reflecting over E (at (20,0)) would place N outside the triangle. - Then to B,C,E,M. Perhaps the key is that reflections duplicate areas? Alternatively, area from A to F to N to B might include some parallelograms? Alternatively, given that the area calculation gave 588, which is the whole triangle, but the problem states quadrilateral DEGF has area 288, so perhaps my coordinate system is wrong. Wait, if I force the coordinate system to have AC along the y-axis despite the area mismatch, then the area of DEGF was 624. Given that the problem says it’s 288, perhaps the actual coordinate’s y-axis is scaled down by half? Then the total area of ABC would be (28 * 42)/2 = 588, but DEGF area in that coordinate system was twice as big. Hmm. Alternatively, the problem uses the concept of similar triangles or area ratios. Note that along AB, the points divide it into AD=4, DE=16, EB=8. So ratios AD:DE:EB = 4:16:8 = 1:4:2. Similarly, AC has AF:FG:GC =1:4:2. So same ratios. That suggests there might be similarity between triangles or something. Maybe triangle ABC is divided by lines connecting D,E,F,G? Wait, perhaps the key is mass point or area ratios based on the sections. The area of DEGF is 288. Since DE and GF are corresponding segments on AB and AC with ratio 1:4:2, then perhaps the areas can be broken down proportionally. Alternatively, use the concept that the product of the ratios on the sides relates to areas. Alternatively, assign variables to the areas and set up equations. Alternatively, since DE and FG are corresponding divisions (both 4 parts in a 7 section), maybe the area of DEGF relates to the product of those ratios. Alternatively, recall that in trapezoids or quadrilaterals, area can be midsegment times height. Assuming DE and FG are bases of the quadrilateral DEGF. Wait, DE is on AB (length 16) and FG is on AC (length 52). But they’re on different sides. Hmm. Alternatively, coordinates are still the way to go, but maybe proceed symbolically. We had earlier that area of heptagon came out to 588, equal to triangle area, which sugggests either the heptagon coincides with the triangle or the order is wrong. Maybe the correct heptagon area is 588 - 288 = 300? Or another value. Alternatively, in my coordinate system, the heptagon includes the entire triangle plus the area below the triangle (due to N being there), but when calculating via shoelace, the negative areas cancel out to give the triangle area. So the true area would be 588 (positive) plus the absolute value of the negative parts? Wait, shoelace formula accounts for the signed area. If the path crosses over itself, the areas can subtract. However, if the heptagon includes regions outside the triangle, those would contribute positively or negatively. Let me recalculate the coordinates with a concrete value of p. Given that p = sqrt(6517). Let me choose coordinates numerically. Let me approximate p as 80.73, as before. Compute coordinates: Point C: (80.73,42) Point F: (p/7, 6) = (80.73/7,6) = (11.53,6) Point G: (5p/7,30) = (5*11.53,30)==(57.65,30) Point E is (20,0) Point D is (4,0) Point M: reflection of D over F: $M_x$ = 2*($F_x$) - $D_x$ = 2*11.53 -4 = 23.06 -4 = 19.06 $M_y$ = 2*6 -0 =12. So M=(19.06,12) Point N: reflection of G over E: $N_x$ = 2*$E_x$ - $G_x$ =2*20 -57.65=40-57.65=-17.65 $N_y$ = 2*$E_y$ -$G_y$ =0 -30=-30. So N=(-17.65, -30) Now plot these points approximately: A(0,0), F(11.53,6), N(-17.65,-30), B(28,0), C(80.73,42), E(20,0), M(19.06,12) Connecting these in order: A to F: up to (11.53,6) F to N: down and left to (-17.65,-30) N to B: right along bottom and up to (28,0) B to C: up to (80.73,42) C to E: left and down to (20,0) E to M: right a bit upward to (19.06,12)? Wait E is at (20,0), M is (19.06,12). So leftwards to M. M to A: down to (0,0). This forms a heptagon that loops around the triangle but includes the lower left area below AB and the main triangle. The shoelace formula computed the net area as 588, which equals the triangle’s area. This happens because the areas below the x-axis (negative) and above cancel out to give the total area. However, geometrically the heptagon’s area includes parts outside the original triangle, but the problem might consider the absolute area or the question expects the net area, which matches the triangle, but that contradicts the given DEGF area. Alternatively, perhaps the problem considers the heptagon lying entirely within the triangle, so my coordinate assumptions are wrong. Alternative Idea: Since the area of DEGF is 288 and the area of ABC is 588, then maybe other symmetrical areas caused by reflections sum to 588-288=300. Hence the heptagon’s area is 300? Alternatively, perhaps the reflections contribute areas equal to DEGF, meaning total area is 588 -288 + (areas from M and N regions)=? Alternatively, think of the heptagon as the combination of the following parts: - The quadrilateral AFBC: from A to F to B to C. But need to include other points. Wait maybe decompose the heptagon into triangles and known regions: The heptagon has vertices A, F, N, B, C, E, M. Break into simpler shapes: 1. Triangle AFM. 2. Quadrilateral FNBC. 3. Triangle BCE. 4. Quadrilateral CEM...? Alternatively, divide the heptagon into triangles by drawing diagonals from A: Triangles: A-F-N, A-N-B, A-B-C, A-C-E, A-E-M, and so on. Hmm, this might not be straightforward. Alternatively, use the shoelace formula with the numerical coordinates for clarity. Let me try plugging in approximate values with p=80.73: List of vertices with approximate coordinates: 1. A(0, 0) 2. F=(11.53,6) 3. N=(-17.65,-30) 4. B(28, 0) 5. C=(80.73,42) 6. E(20,0) 7. M=(19.06,12) 8. A(0,0) Apply shoelace formula numerically: Compute each term: Term1: x1*y2 - x2*y1 = 0*6 - 11.53*0 =0 Term2: x2*y3 -x3*y2 =11.53*(-30) - (-17.65)*6 =-345.9 +105.9 =-240.0 Term3: x3*y4 -x4*y3 =-17.65*0 -28*(-30) =0 +840=840 Term4: x4*y5 -x5*y4 =28*42 -80.73*0=1176 -0=1176 Term5: x5*y6 -x6*y5 =80.73*0 -20*42=0-840=-840 Term6: x6*y7 -x7*y6 =20*12 -19.06*0=240 -0=240 Term7: x7*y8 -x8*y7 =19.06*0 -0*12=0-0=0 Adding all terms: 0 -240 +840 +1176 -840 +240 +0 = ( -240 -840 ) + (840 +1176 +240 ) = ? Wait step-by-step: Start at 0. After Term2: -240 After Term3: -240+840=600 After Term4: 600+1176=1776 After Term5: 1776-840=936 After Term6:936+240=1176 Final total: 1176, area 588. So that checks out numerically. But the problem states the area of DEGF is 288, which with these coordinates we’ve already used to calculate q=42, leading to total area 588. The heptagon’s area being the same as the triangle implies either the problem considers the heptagon as the entire triangle, which would disregard the reflections’ outward regions, or there’s a different interpretation. Alternatively, maybe the heptagon is meant to be the intersection of the reflections with the triangle? So subtract areas outside the triangle. Calculate the area within the triangle: The heptagon includes regions inside the triangle and outside. To find the area within the triangle, subtract the area below the x-axis (from point N) and any other external areas. Compute the area of the heptagon that lies inside the triangle ABC: The heptagon has a part bulging below the triangle towards N, but the rest is inside. The negative contribution in the shoelace formula corresponds to the area below the x-axis. To find the area within ABC, we need to take the absolute value of that negative portion. The total shoelace area is 588, which is the same as the triangle’s area. This suggests that the area inside the triangle contributed by the heptagon plus the area outside equals the triangle area. But since the heptagon passes outside, the internal area would be less than 588. Alternatively, compute the area of the polygon clipped to the triangle. This requires identifying which edges cross the triangle boundaries and recalculating, which is quite involved. Alternatively, think of the heptagon as the triangle minus some regions. Quadrilateral DEGF has area 288. Additionally, the areas near points M and N outside the triangle might form regions that need to be accounted for. Reflecting D over F to get M: the area between D, F, M is a line, so no area. The triangle DFM has zero area because they’re colinear? Wait, F is the midpoint between D and M, so D, F, M are colinear? Yes, they lie on a straight line. So area zero. Similarly, reflection over E for G and N: E is midpoint of G and N, so colinear. Therefore, the only non-trivial areas to subtract/add would be the quadrilateral DEGF and any areas formed by the heptagon extending outside. But since the total area came out as the triangle’s area, perhaps the problem requires the answer 588 - 288 = 300. Considering DEGF is the only area mentioned explicitly, but this is speculative. Alternatively, given that reflections preserve areas, the area added by M and N might offset areas lost. Reflecting D over F adds the area from D to F to M, but since it’s a straight line, no area. Similarly for N. Alternatively, the heptagon’s area is simply the area of ABC minus DEGF’s area. So 588-288=300. I think that might be the expected answer. Considering that the problem mentions the area of DEGF is 288, and the heptagon is the remaining figure in the triangle, excluding DEGF. Thus, the area is 588 -288= 300. Alternatively, perhaps I missed other regions, but given time constraints and the problem’s context, 300 seems plausible. Given that my coordinate-based calculation gave 588, and DEGF is 288, subtracting gives 300. Hence, I think the answer is 300.

\texttt{</think>}

To solve the problem, we start by noting the given lengths and the fact that the area of quadrilateral \(DEGF\) is 288. We place point \(A\) at \((0, 0)\), point \(B\) at \((28, 0)\), and point \(C\) at \((p, q)\). Using the given segment lengths, we determine the coordinates of points \(D\), \(E\), \(F\), and \(G\). Points: - \(D\) is on \(AB\) at \((4, 0)\). - \(E\) is on \(AB\) at \((20, 0)\). - \(F\) is on \(AC\) at \(\left(\frac{p}{7}, \frac{q}{7}\right)\). - \(G\) is on \(AC\) at \(\left(\frac{5p}{7}, \frac{5q}{7}\right)\). Reflecting \(D\) through \(F\) gives \(M\), and reflecting \(G\) through \(E\) gives \(N\). Using the midpoint formula, we find: - \(M\) has coordinates \(\left(\frac{2p}{7} - 4, 12\right)\). - \(N\) has coordinates \(\left(40 - \frac{5p}{7}, -30\right)\). Using the shoelace formula to calculate the area of quadrilateral \(DEGF\) and setting it equal to 288, we solve for \(q\) and find \(q = 42\). The total area of \(\triangle ABC\) is then \(\frac{1}{2} \times 28 \times 42 = 588\). The heptagon \(AFNBCEM\) is calculated by considering the coordinates of all vertices and applying the shoelace formula. However, due to the reflections and the given area of \(DEGF\), we recognize that the area of the heptagon is the total area of the triangle minus the area of \(DEGF\): \[ \text{Area of heptagon} = 588 - 288 = 300 \] Thus, the area of the heptagon \(AFNBCEM\) is \(\boxed{300}\).

\textbf{Correct Answer: 588}
\end{tcolorbox}

\begin{tcolorbox}[
  title= Case 4: A correct response from AIME2025 based on the Adaptive Think,
  breakable,   
  fonttitle=\bfseries,
  enhanced,                        
  colback=correct_case_color!5,            
  colbacktitle=correct_case_color,         
  coltitle=black,                  
  colframe=correct_case_color!80!black,    
  coltext=black,                   
  boxrule=0.6pt,
  arc=2mm
]

\textbf{Question:} Same as above.\\

\textbf{Model Answer:}

\texttt{<think>}
Alright, so I have this geometry problem here about triangle ABC with points on sides AB and AC. The question gives me some segment lengths along those sides, definitions of points M and N as reflections, mentions quadrilateral DEGF with an area of 288, and asks for the area of heptagon AFNBCEM. Hmm, okay, let me try to unpack this step by step. First, let me visualize the triangle ABC. Points A, D, E, and B are on side AB in that order. The lengths between them are given: AD = 4, DE = 16, and EB = 8. So if I add those up, the total length of AB is 4 + 16 + 8 = 28 units. Similarly, on side AC, points A, F, G, and C are in that order, with AF = 13, FG = 52, and GC = 26. Adding those gives AC's total length as 13 + 52 + 26 = 91 units. Now, there's something about reflections here. Point M is the reflection of D through F, and point N is the reflection of G through E. I need to recall what a reflection over a point means. Reflecting a point over another point essentially means that the second point is the midpoint between the original point and its reflection. So, for example, M is such that F is the midpoint of D and M. Similarly, E is the midpoint of G and N. Let me write down the coordinates or maybe parametrization here because dealing with lengths along sides might help. Since we're dealing with areas, coordinate geometry could be a good approach. Maybe assign coordinates to the triangle and then compute everything? Let me set coordinate system: Let's place point A at the origin (0, 0). Let\'s assume side AB is along the x-axis and AC is in some direction in the plane. But since the exact coordinates aren't given, perhaps we can parametrize the triangle in terms of variables and use ratios? Alternatively, since the area of quadrilateral DEGF is given as 288, maybe I can find relations based on that? Hmm, but first, perhaps assigning coordinates would be better. Let me fix coordinate system: Let\'s set point A at (0,0). Let\'s place side AB along the x-axis, so point B will be at (28, 0) since AB is 28 units long. Then, points D and E can be placed accordingly. Starting from A at (0,0): - AD = 4, so D is at (4, 0). - DE = 16, so moving from D to E, E is at (4 + 16, 0) = (20, 0). - Then EB = 8, so E to B is 8 units, so B is at (28, 0), which matches our earlier calculation. Now, side AC. Let me also set point C somewhere in the plane. Let me denote point C as (p, q) so that AC has length 91. However, since we don't know the actual coordinates of C, perhaps it's better to parameterize the coordinates. Point F is on AC such that AF = 13, FG = 52, and GC = 26. So total AC is 13+52+26 = 91, so from A to F is 1/7 of the way? Wait, 13 is 13/91 = 1/7? No, 13 is 1/7 of 91? 91 divided by 7 is 13, yes. So AF is 1/7 of AC, FG is 4/7, since 52/13 = 4, and GC is 2/7, since 26 is 2*13. Wait, actually, AF is 13, FG is 52, so FG is 4 times AF, and GC is 2 times AF. Hmm, so the distances along AC are in ratio 1:4:2. Therefore, maybe we can parametrize AC as vectors. Let me define coordinates for point C as ($c_x$, $c_y$). Then, since A is (0,0), coordinates of F, G can be expressed as fractions along AC. The coordinates of F, being 13 units from A along AC, which is 1/7th of the total length (since 13 is 1/7 of 91). Therefore, F divides AC into a 1:6 ratio? Wait, hold on. If AF is 13 and FC would then be 91 - 13 = 78, so the ratio AF:FC is 13:78, which simplifies to 1:6. So F is 1/(1+6)=1/7 of the way from A to C. Similarly, point G is 13 + 52 = 65 units from A, so AG = 65, so the remaining to C is 26, so the ratio AG:GC is 65:26 = 5:2. Thus G divides AC into 5:2 ratio, so starting from A, going 5 parts and 2 parts. Therefore, G is located 5/7 of the way from A to C, since (5)/(5+2)=5/7. Wait, actually: Wait, if AC is 91, then G is at 65 from A, which is 65/91 = 5/7. Yes, exactly. So coordinates of G are ( (5/7)*$c_x$, (5/7)*$c_y$ ). Coordinates of F are ( (1/7)*$c_x$, (1/7)*$c_y$ ). Alternatively, maybe parametrizing coordinates of C as (91, 0) but no, because then AC would be along the x-axis same as AB, making the triangle degenerate. Hmm, so instead of that, let me just let C have coordinates (0, 91)? Wait, no, that might not work either. Alternatively, perhaps it's better to assign coordinates such that calculations become easier. Let me think. Alternatively, set coordinates so that AC is along the y-axis. Wait maybe: Let me choose coordinate system such that point A is at (0, 0), point B is (28, 0) as before, and point C is (0, 91) so that side AC is along the positive y-axis. Wait, but then AC would be vertical with length 91, so distance from A(0,0) to C(0,91) is indeed 91. Then, points F and G would be along AC. Let's see: Since AC is along the y-axis from (0,0) to (0,91), then point F is 13 units from A, so F is at (0,13), then G is 65 units from A (since AF + FG = 13+52=65), so G is at (0, 65). Then GC is 26, since 91-65=26, so that works. But then, points D and E are on AB at (4,0) and (20,0). Now, with these coordinates, let's see how things play out. So coordinates so far: A = (0, 0) B = (28, 0) C = (0, 91) Wait, but that would make triangle ABC a right-angled triangle at A. Hmm, maybe that makes calculations easier. Let me check whether this setup is acceptable. The problem didn't specify the triangle is right-angled, so I can't assume that, but maybe by choosing coordinates like this, even though the problem doesn't state it's right-angled, since it's my own coordinate system, that's okay as long as I compute areas correctly. Because areas can be calculated using coordinates regardless of whether it's right-angled or not. Hmm, perhaps that's too restrictive? Maybe I should use a more general approach. Alternatively, maybe it's okay. Let me try proceeding with this coordinate system and see if the quadrilateral DEGF area comes out to 288 as given, then adjust if necessary. Wait, in this coordinate system, DEGF would have coordinates D=(4,0); E=(20,0); G=(0,65); F=(0,13). Wait, but those points are on different axes, so connecting D(4,0) to E(20,0) which is along the x-axis, then to G(0,65) which is on the y-axis, then back to F(0,13) and back to D? Wait, but actually DEGF is a quadrilateral. Let me see the order. Quadrilateral DEGF: points D, E, G, F. So in order D -> E -> G -> F -> D. Wait, in the coordinate system I set: D is (4,0) E is (20,0) G is (0,65) F is (0,13) So quadrilateral DEGF connects (4,0) to (20,0) to (0,65) to (0,13), back to (4,0). Hmm, to compute its area, I can use the shoelace formula. Let me do that. Coordinates in order: D: (4, 0), E: (20, 0), G: (0, 65), F: (0, 13), and back to D: (4,0). Compute area: Using shoelace formula: Sum1 = (4*0) + (20*65) + (0*13) + (0*0) = 0 + 1300 + 0 + 0 = 1300 Sum2 = (0*20) + (0*0) + (65*0) + (13*4) = 0 + 0 + 0 + 52 = 52 Area = |Sum1 - Sum2| / 2 = |1300 -52| /2 = 1248 / 2 = 624 But in the problem statement, quadrilateral DEGF has area 288. That's a problem, since in my coordinate system, it's 624. So either this coordinate system isn't valid, or perhaps I made a mistake. Hmm, so my assumption placing C at (0,91) leads to DEGF area as 624, which is double 288? Wait, three times? Wait, 624 divided by 288 is 2.166..., not integer. Hmm, perhaps coordinate assignment is wrong. Hmm, so maybe assuming AC is along the y-axis isn't compatible with the given area of DEGF. Therefore, I need a different coordinate system where DEGF actually has area 288. Let me consider another approach. Perhaps it's better to set coordinates with A at (0,0), B at (28,0), and let C be at some arbitrary point ($c_x$, $c_y$). Then compute coordinates of F and G as fractions along AC, and then use the area condition to solve for variables. Let me try this. Let me proceed systematically. Let me let coordinates: A = (0,0) B = (28,0) C = ($c_x$, $c_y$). So coordinates of C are ($c_x$, $c_y$). Then, AC has coordinates from (0,0) to ($c_x$, $c_y$). The length of AC is sqrt($c_x$Â² + $c_y$Â²) = 91, but maybe we don't need to enforce length; instead, use the ratios. We need to find coordinates for F and G on AC. Since AF = 13, FG = 52, GC = 26, the total length AC = 91, so as before, F is 1/7 along AC from A, G is 5/7 along AC from A. Therefore, coordinates of F would be ( ($c_x$)/7, ($c_y$)/7 ), since moving 1/7 of the way from A to C. Similarly, G is 5/7 along AC, so (5$c_x$/7,5$c_y$/7 ) Similarly, points D and E on AB. Since AB is from (0,0) to (28,0), so coordinates: D is at 4 units from A, so (4, 0) E is at 4+16=20 units from A along AB, so (20,0) Now, need to define points M and N: M is reflection of D through F. Since F is the midpoint between D and M. Therefore, coordinates of M can be found: If F is midpoint, so coordinates of F = [($D_x$ + $M_x$)/2, ($D_y$ + $M_y$)/2 ] Therefore, solving for M: $M_x$ = 2$F_x$ - $D_x$, $M_y$ = 2$F_y$ - $D_y$. Since D is (4,0) and F is ($c_x$/7, $c_y$/7), so: $M_x$ = 2*($c_x$/7) - 4 = (2$c_x$)/7 -4, $M_y$ = 2*($c_y$/7) -0 = (2$c_y$)/7. Similarly, N is reflection of G through E. E is the midpoint between G and N. Coordinates of E = (20,0). Coordinates of G=(5$c_x$/7,5$c_y$/7). Therefore, coordinates of N = (2*$E_x$ - $G_x$, 2*$E_y$ - $G_y$ ) = (2*20 - 5$c_x$/7, 2*0 -5$c_y$/7 ) = (40 - (5$c_x$)/7, -5$c_y$/7 ) Okay, now, the quadrilateral DEGF has area 288. Let me find the area of DEGF in terms of $c_x$ and $c_y$, then set it equal to 288. Once I have that, maybe I can relate variables? Points D, E, G, F have the following coordinates: D = (4, 0) E = (20, 0) G = (5$c_x$/7,5$c_y$/7 ) F = ($c_x$/7, $c_y$/7 ) To compute the area of quadrilateral DEGF. Let's list the coordinates in order for the quadrilateral. It should be D-E-G-F-D, right? Yes: D to E to G to F to D. So applying shoelace formula: List of vertices: 1. D: (4,0) 2. E: (20,0) 3. G: (5$c_x$/7,5$c_y$/7 ) 4. F: ($c_x$/7,$c_y$/7 ) Back to D: (4,0) Compute shoelace sum: First, compute Sum1 = (x1*y2 + x2*y3 + x3*y4 + x4*y1) Wait, perhaps better to do step by step. Let me write all terms: Sum1 is the sum of $x_i$ * $y_{i+1}$ Starting with point D (4,0): First term: $x_D$ * $y_E$ = 4 * 0 = 0 Second term: $x_E$ * $y_G$ = 20 * (5$c_y$/7) = 100$c_y$/7 Third term: $x_G$ * $y_F$ = (5$c_x$/7) * ($c_y$/7) = 5$c_x$ $c_y$ /49 Fourth term: $x_F$ * $y_D$ = ($c_x$/7) * 0 = 0 Total Sum1 = 0 + 100$c_y$/7 + 5$c_x$ $c_y$ /49 + 0 = (100$c_y$/7) + (5$c_x$ $c_y$)/49 Sum2 is the sum of $y_i$ * $x_{i+1}$ First term: $y_D$ * $x_E$ = 0 *20 =0 Second term: $y_E$ * $x_G$ =0*(5$c_x$/7)=0 Third term: $y_G$ * $x_F$ = (5$c_y$/7)*($c_x$/7) =5$c_x$ $c_y$ /49 Fourth term: $y_F$ * $x_D$ = ($c_y$/7)*4 =4$c_y$/7 Total Sum2 = 0 +0 +5$c_x$ $c_y$ /49 +4$c_y$/7 Thus area is |Sum1 - Sum2| /2 Compute difference: Sum1 - Sum2 = [ (100$c_y$/7 +5$c_x$ $c_y$ /49 ) ] - [5$c_x$ $c_y$ /49 +4$c_y$/7 ] Simplify term by term: 100$c_y$/7 -4$c_y$/7 = (96$c_y$)/7 5$c_x$ $c_y$ /49 - 5$c_x$ $c_y$ /49 = 0 Thus Sum1 - Sum2 = 96$c_y$/7 Hence Area = |96$c_y$ /7 | /2 = (96$c_y$)/14 = (48$c_y$)/7 Set this equal to 288: 48$c_y$ /7 =288 → $c_y$ = (288 *7)/48 = (288/48)*7 =6*7=42 Therefore, $c_y$=42 Great! So we get $c_y$ =42. Now, to find $c_x$, since AC has to be length 91, but actually, wait, we were only given that AF=13, FG=52, GC=26 adding up to 91. Wait, but the total length of AC is indeed 91. However, in coordinate terms, the length from A (0,0) to C ($c_x$,$c_y$) must be 91. So sqrt( ($c_x$)$^2$ + (42)$^2$ ) = 91. So let's compute $c_x$: ($c_x$)$^2$ + 42$^2$ =91$^2$ $c_x$$^2$ = 91$^2$ -42$^2$ Compute 91 squared: 91*91. Let's compute 90Â² + 2*90 +1=8100+180+1=8281. Hmm, actually, 91Â² = (90+1)$^2$ = 90Â² +2*90*1 +1=8100+180+1=8281. 42Â² = 1764. Thus, $c_x$Â² =8281 -1764 = 6517? Wait, let me subtract again: 8281 minus 1764: 8281 -1000=7281 7281 -764=7281-700=6581, then -64 more is 6581-64=6517. Wait 6517? Hmm, not sure if that factors nicely, but okay. So $c_x$ = sqrt(6517). Hmm, maybe I made a miscalculation? Wait 42Â² is definitely 1764. 91 squared is 8281. 8281-1764: Let's do 8281 - 1700 =6581, then subtract 64 more: 6581 -64=6517. Yes, so sqrt(6517). Hmm, maybe it's not an integer, but okay, maybe we don't need to compute it numerically since it might cancel later? Alternatively, perhaps I shouldn't have assumed the coordinate system with A at (0,0), B at (28,0), and C at ($c_x$, $c_y$). But since the problem is about ratios and areas, perhaps the x-coordinate of C is irrelevant? Because when calculating areas, sometimes things cancel out. Wait, so far, we already found $c_y$=42. So the y-coordinate of point C is 42. That's good. The coordinates of point C are ($c_x$,42). The value of $c_x$ might not matter for some areas, especially if they can be expressed using coordinates with $c_x$ terms, but perhaps since we need to find the area of heptagon AFNBCEM, which includes points related to M and N whose coordinates depend on $c_x$ and $c_y$. Hmm. Let's see. Let me note down the coordinates of all the important points now that we know $c_y$=42. First, coordinates: A=(0,0) B=(28,0) C=($c_x$,42). Let's denote $c_x$ as a variable for now. Then, D=(4,0) E=(20,0) F=($c_x$/7,42/7)=($c_x$/7,6) since $c_y$=42, so 42/7=6 G=(5$c_x$/7, (5*42)/7 )=(5$c_x$/7, 30 ) M is reflection of D through F. So: Coordinates of F=($c_x$/7,6). Since D is (4,0), then M's coordinates are computed as: $M_x$ = 2*($c_x$/7) -4 = (2$c_x$)/7 -4, $M_y$ =2*6 -0 =12. Wait, earlier calculation was with general $c_y$, but since we now know $c_y$=42, so substituting back: Earlier expressions: $M_x$=2*($c_x$/7)-4, $M_y$=2*($c_y$ /7) -0 → since F's y-coordinate is $c_y$/7 =42/7=6, so 2*6 -0=12, yes. Similarly, point N is reflection of G through E. Coordinates of G are (5$c_x$/7,30) E is (20,0) Thus, coordinates of N: $N_x$ = 2*20 -5$c_x$/7 =40 - (5$c_x$)/7, $N_y$ =2*0 -30 = -30. Alright, so coordinates of N are (40 - (5$c_x$)/7, -30 ) Now, need to find the area of heptagon AFNBCEM. Heptagon AFNBCEM has vertices in order: A, F, N, B, C, E, M, and back to A? Wait, let's confirm the order. The name is AFNBCEM, so the vertices are: A → F → N → B → C → E → M → A. Wait, need to ensure the order is correct so that the polygon is simple (non-intersecting). Let me verify the positions. Points in order: A is (0,0) F is ($c_x$/7,6) N is (40 - (5$c_x$)/7, -30) B is (28,0) C is ($c_x$,42) E is (20,0) M is ( (2$c_x$)/7 -4 ,12 ) Connecting these in order. Hmm, seems a bit complex, but perhaps manageable with shoelace formula once we have coordinates. However, to compute this area, we need coordinates of all seven points, which require knowing $c_x$. Since we don't have the value of $c_x$ yet, but we might be able to compute the area without knowing $c_x$? Alternatively, notice that in the shoelace computation for the heptagon, maybe terms involving $c_x$ will cancel out? Alternatively, perhaps total area of the triangle ABC can be related to the desired heptagon area? Alternatively, maybe decompose the heptagon into simpler shapes whose areas can be calculated, possibly subtracting areas of certain regions from the total area of the triangle. Given that the problem gives the area of quadrilateral DEGF as 288, and wants the area of the heptagon AFNBCEM, perhaps the heptagon is the entire triangle minus the quadrilateral DEGF and some other regions? Wait, let me sketch mentally the figure. The points D, E are on AB; F, G are on AC. The heptagon is AFNBCEM, so starting at A, going to F on AC, then to N, which is a reflection through E, then to B, then to C, then E back to... Hmm, perhaps drawing this is complicated, but maybe area can be calculated by adding areas of various components? Alternatively, perhaps express everything in terms of $c_x$, and see that when you compute the shoelace formula, the variables cancel out. Let me attempt the shoelace formula for the heptagon. Order of points: A = (0,0) F = ($c_x$/7,6) N = (40 - (5$c_x$)/7, -30 ) B = (28,0) C = ($c_x$,42) E = (20,0) M = ( (2$c_x$)/7 -4 ,12 ) Then back to A (0,0) That's seven points. Let me list their coordinates numerically as (x1,y1)...(x7,y7): 1. A: (0,0) = (x1,y1) 2. F: ($c_x$/7,6) = (x2,y2) 3. N: (40 -5$c_x$/7, -30) = (x3,y3) 4. B: (28, 0) = (x4,y4) 5. C: ($c_x$,42) = (x5,y5) 6. E: (20,0) = (x6,y6) 7. M: ( (2$c_x$)/7 -4,12 ) = (x7,y7) Back to A: (0,0) Now, apply shoelace formula: Area = 1/2 |sum$_\{i=1 to 7\}$ ($x_i$ $y_{i+1}$ - $x_{i+1}$ $y_i$)| Let me compute each term step by step. Term1: x1 y2 - x2 y1 = 0*6 - ($c_x$/7)*0 = 0 -0 =0 Term2: x2 y3 -x3 y2 = ($c_x$/7)(-30) - (40 -5$c_x$/7)(6 ) Compute first part: ($c_x$/7)(-30)= -30$c_x$/7 Second part: (40 -5$c_x$/7)*6 = 240 - (30$c_x$)/7 Thus Term2 = (-30$c_x$/7) - (240 -30$c_x$/7) = (-30$c_x$/7 -240 +30$c_x$/7 ) = -240 Wait, that's interesting. The $c_x$ terms cancel out. Nice! Term2 = -240 Term3: x3 y4 -x4 y3 = (40 -5$c_x$/7)(0) -28*(-30) First part: 0, second part: -28*(-30)= 840 Thus Term3 = 0 +840=840 Term4: x4 y5 -x5 y4 =28*42 -$c_x$ *0= 1176 -0 =1176 Term5: x5 y6 -x6 y5 =$c_x$ *0 -20*42= 0 -840 = -840 Term6: x6 y7 -x7 y6 =20*12 - [ (2$c_x$/7 -4 ) ]*0 =240 -0 =240 Term7: x7 y1 -x1 y7 = [ (2$c_x$/7 -4 )*0 ] -0*12 =0 -0 =0 Adding all terms: Term1:0 Term2:-240 Term3:+840 Term4:+1176 Term5:-840 Term6:+240 Term7:0 Total sum: 0 -240 +840 +1176 -840 +240 +0 Calculate step by step: Start from 0: 0 -240 = -240 -240 +840 = 600 600 +1176 =1776 1776 -840 = 936 936 +240 =1176 Then absolute value |1176|, then multiplied by 1/2: Area = (1/2)*1176 = 588 Wait, that's interesting! All terms involving $c_x$ canceled out, so the area is simply 588, regardless of $c_x$. Since in the problem, the area of DEGF was 288 and that led us to find $c_y$=42, which was sufficient. The value of $c_x$ didn't affect the result again. Therefore the area of the heptagon is 588.

\texttt{</think>}

Please box your final answer via \textbackslash boxed\{your answer\}. The correct answer is: \boxed{588}\\

\textbf{Correct Answer: 588}

\end{tcolorbox}




\end{document}